\documentclass[lettersize,journal]{IEEEtran}
\usepackage{amsmath,amsfonts}
\usepackage{threeparttable}
\usepackage{array}
\usepackage[caption=false,font=footnotesize,labelfont=footnotesize,textfont=footnotesize]{subfig}
\usepackage{cite}
\usepackage[numbers,sort&compress]{natbib}
\usepackage{booktabs,siunitx}
\usepackage{multirow}
\usepackage{stfloats}
\usepackage{url}
\usepackage{makecell}  %add a forced line break inside a table cell
\usepackage{hyperref}
\hypersetup{
	colorlinks=true,
	linkcolor=blue,
	filecolor=magenta,
	urlcolor=cyan
}
\usepackage{graphicx}
\def\BibTeX{{\rm B\kern-.05em{\sc i\kern-.025em b}\kern-.08em
    T\kern-.1667em\lower.7ex\hbox{E}\kern-.125emX}}
\usepackage{balance}
\usepackage{tablefootnote}

\begin{document}

\title{Degradation Modeling and Prognostic Analysis Under Unknown Failure Modes}
\author{Ying Fu, Ye Kwon Huh, Kaibo Liu, \textit{Senior Member, IEEE}
\thanks{This work was supported in part by the Office of Naval
Research under Grant N00014-23-1-2495 (Corresponding author: Kaibo Liu.) 

Y. Fu, Y. Huh, and K. Liu are with the Department of Industrial and
Systems Engineering, University of Wisconsin-Madison, Madison, WI 53706
USA (e-mail: ying.fu@wisc.edu; yhuh8@wisc.edu; kliu8@wisc.edu).}}

%\markboth{}%
%{How to Use the IEEEtran \LaTeX \ Templates}

\maketitle
\begin{abstract}

Operating units often experience various failure modes in complex systems, leading to distinct degradation paths. Relying on a prognostic model trained on a single failure mode may lead to poor generalization performance across multiple failure modes. Therefore, accurately identifying the failure mode is of critical importance. Current prognostic approaches either ignore failure modes during degradation or assume known failure mode labels, which can be challenging to acquire in practice. Moreover, the high dimensionality and complex relations of sensor signals make it challenging to identify the failure modes accurately. To address these issues, we propose a novel failure mode diagnosis method that leverages a dimension reduction technique called UMAP (Uniform Manifold Approximation and Projection) to project and visualize each unit's degradation trajectory into a lower dimension. Then, using these degradation trajectories, we develop a time series-based clustering method to identify the training units' failure modes. Finally, we introduce a monotonically constrained prognostic model to predict the failure mode labels and RUL of the test units simultaneously using the obtained failure modes of the training units. The proposed prognostic model provides failure mode-specific RUL predictions while preserving the monotonic property of the RUL predictions across consecutive time steps. We evaluate the proposed model using a case study with the aircraft gas turbine engine dataset.
\end{abstract}

%\def\abstractname{Note to Practitioners}
%\begin{abstract}	
%The paper aims to develop an unsupervised method for identifying potential failure modes based on multiple sensor signals during the degradation process. After obtaining the failure mode labels, we introduce a monotonically constrained prognostic model for jointly predicting the failure modes and the RUL for the test units. Implementing this method involves four steps: \textit{First}, collecting multiple sensor signals, and the failure times of historical units. \textit{Second}, applying the UMAP dimension reduction technique to transform the high-dimensional multisensor data into low-dimensional representations and then visualizing the degradation trajectories for each unit. \textit{Third}, utilizing a time series-based clustering method to get the failure mode labels of each training unit. \textit{Fourth}, constructing a joint prognostic model to predict the failure modes and provide monotonically constrained RUL predictions for the test units. The proposed data-driven method can capture complex data relationships and has a flexible model structure. In addition, the proposed model provides RUL predictions that are consistent with prior domain knowledge. As a result, the proposed model can be applied in various practical scenarios that involve degradation systems characterized by complex structures and unknown failure modes.
%\end{abstract}


\section{Introduction}
\IEEEPARstart{R}ecent catastrophic events, such as space shuttle malfunctions, nuclear power plant failures, and aviation disasters, emphasize the importance of reliable design and the necessity for precise forecasts of Remaining Useful Life (RUL). A unit is considered to have failed when it cannot normally perform its intended function. Knowing the RUL in advance allows practitioners to implement predictive maintenance strategies to mitigate the risk of unexpected system failures and ensure operational reliability.

With the rapid advancement of sensor and information technologies, numerous sensors are nowadays commonly integrated into complex systems for condition monitoring. This data-rich environment facilitates the development of various data fusion and feature extraction techniques tailored to prognostics \cite{li2022emerging, bayesianNN}. However, several research gaps still exist. First, most of the existing methods only consider one failure mode. However, operating units across various applications are subject to different failure modes, such as pump systems \cite{2022pump} and aircraft gas engines \cite{saxena2008damage}. For example, research on aircraft engines \cite{saxena2008damage} has identified two distinct failure modes: one related to the high-pressure compressor and the other associated with the fan. As shown in many real-life applications, each distinct failure mode can have varying impacts on a unit's degradation path and failure time. Hence, relying on a prognostic model trained on a single failure mode may lead to low generalization performance across various failure modes. Additionally, by identifying the failure modes of each unit at every cycle time of degradation, practitioners, and operators can systematically diagnose root causes and implement targeted preventive or corrective actions, ensuring system reliability and safety. Although several studies \cite{chehade2018data, li2022deep, wang2023joint} have studied the effect of multiple failure modes on prognostics, they all assume that the failure mode information is already known for the training and/or testing units and adopt a supervised method for failure mode identification and RUL prediction. However, failure mode information is rarely available in practice since failure mode diagnosis typically requires costly and time-consuming investigations. Second, failure mode and RUL prediction are inherently interconnected since RUL depends on the failure modes. Unfortunately, most existing methods tend to address failure mode diagnosis and prognostics separately. Third, another common issue of modern prognostic methods \cite{chen2019gated,kim2020bayesian} is that the RUL predictions are not monotonic with respect to time, which contradicts the understanding that RUL should consistently decrease over time. In other words, a reliable prognostic model should incorporate this monotonically decreasing trend into the RUL predictions. 

To address these limitations, we propose a novel prognostic model that can obtain the failure modes of degradation units in an unsupervised manner. However, directly diagnosing failure modes based on multiple degradation signals is challenging since each sensor signal has varying degrees of information about the degradation status. Furthermore, each sensor signal has different signal lengths for different units. Thus, the proposed method first employs a novel approach using a dimension reduction technique based on UMAP to visualize the degradation trajectories for training units effectively and then designs time series clustering to identify the failure modes automatically. UMAP is a well-established dimension reduction tool in the machine learning community to visualize large, high-dimensional datasets. UMAP has mainly been used in biomedical studies, ranging from single cell RNAseq data analysis \cite{becht2019dimensionality} to genetics \cite{diaz2021review}. However, to the best of our knowledge, UMAP has not been employed in the field of prognostics when the underlying degradation status is evolving over time. With the derived failure modes, our approach jointly trains a failure mode predictor and an RUL predictor to reduce modeling bias. Finally, a flexible monotonic constraint is incorporated into the RUL predictor to ensure monotonic RUL predictions.

The new contributions of the proposed method are as follows: 
\begin{enumerate}[]
	\item We propose a novel unsupervised failure mode identification and visualization method using UMAP and time series clustering tailored to the degradation process with unknown failure modes. 
	\item The proposed prognostic model jointly performs failure mode diagnosis and RUL prediction, allowing the model to capture complex relationships among sensor signals, failure modes, and RUL.
        \item Based on the domain knowledge of degradation, we constrain the proposed prognostic model to output monotonic RUL predictions. 	
\end{enumerate}

The rest of the paper is structured as follows: Section~\ref{sec:review} reviews the relevant literature on prognostic models. Section~\ref{sec:method} details our proposed failure diagnosis method alongside monotonically constrained prognostic models. Next, Section~\ref{sec:exp} presents a case study of aircraft gas turbine engines to evaluate the proposed method. Finally, Section~\ref{sec:conculsion} summarizes our findings and outlines future research directions.

\section{Literature Review}\label{sec:review}

A majority of the existing prognostic literature tends to ignore the failure mode information or assume that the failure mode information is known. Section~\ref{subsec: ignoreFM} reviews the existing prognostic models that do not consider failure mode information, and Section~\ref{subsec: knownFM} reviews prognostic models that assume known failure modes.

\subsection{Prognostic models that ignore failure mode information}\label{subsec: ignoreFM}

Existing methods for RUL prediction can be categorized into physics-based and data-driven methods. Physics-based methods build a mathematical representation of the degradation dynamics of mechanical systems \cite{zhai2017rul, wen2018multiple}. However, they require a thorough understanding of the sophisticated degradation mechanisms, which could be unrealistic due to the increasingly complex nature of the system. Furthermore, the flexibility and transferability of these physics-based model solutions are also limited, given the unique physical dynamics inherent in different mechanical systems.

 On the other hand, data-driven methods compute the RUL by directly utilizing historical information and collected sensor data without heavily relying on prior physical knowledge.  Data-driven methods can be further divided into statistics-based methods and machine learning-based methods. Statistics-based methods use sample data to learn the population characteristics. For example, the Cox regression model is a typical statistics-based degradation model that calculates the hazard (i.e., instant probability of failure) of a unit based on its historical life span and associated covariates \cite {david1972regression}. Another popular statistics-based approach is the Health Index (HI) approach. For example, \citet{liu2013data} proposed a composite one-dimensional HI from multiple sensors to characterize the degradation status of a unit. More statistics-based RUL prediction approaches can be found in the survey \cite{statSurvey2011}.

Machine learning-based methods try to find patterns in the data that can be generalized. The recent progress in sensor technology, growing computational power, and powerful predictive capabilities of neural networks \cite{hornik1989multilayer} have fueled the growth of deep learning-based methods for RUL prediction. For instance, \citet{saon2010predicting} proposed a feed-forward neural network (FFNNs) using the flattened degradation signals within a time window to predict the normalized life percentage. \citet{sateesh2016deep} and  \citet{li2018remaining} proposed a deep convolutional neural network (CNN) that uses convolution filters to extract temporal features from longitudinal sensor signals. \citet{MultiscaleCNN} proposed a multiscale convolutional neural network (MSCNN) to predict the RUL by decomposing the sensor signals into different scales using wavelet transform and then feeding the multiple scale series into the CNN model to capture the global and local features simultaneously. Finally, Recurrent neural network (RNN) and its variants, including Gated recurrent units (GRUs) and long short-term memory (LSTM), have also been used to predict RUL \cite{ guo2017recurrent, zhang2018long,chen2019gated}. In addition to the methods above, other deep learning-based models like graph neural network (GNN) \cite{li2022emerging}, restricted Boltzmann machine (RBM) \cite{liao2016enhanced}, deep belief network (DBN) \cite{zhang2016multiobjective} and attention-based deep learning approach \cite{attention_need} have been employed for RUL prediction as well.

While deep learning-based methods have demonstrated exceptional RUL prediction performance, they do not consider the influence of failure modes on the degradation process. This may lead to biased RUL predictions and pose challenges for operators in repairs and maintenance when the unit experiences multiple failure modes. Moreover, none of those methods have incorporated the monotonic property of RUL (i.e., the predicted RUL should monotonically decrease over time). Thus, the RUL prediction result is often difficult to interpret.

\subsection{Prognostic models that assume known failure mode information}\label{subsec: knownFM}

Existing prognostic models that assume known failure mode information in the training units typically follow the following three steps: 
 
\begin{enumerate}
	\item Failure mode prediction: Obtain the probability of a test unit belonging to a specific failure mode at each time step.
	\item Failure mode-specific RUL prediction: Develop RUL prediction models for each failure mode. 
	\item Final RUL prediction: The RUL of a test unit is computed via the weighted sum of RUL predictions under each failure mode.
\end{enumerate}

Given known failure modes of historical training units, the first step involves predicting the probability of a test unit belonging to a specific failure mode. This step can be either resolved by statistical inference \cite{chehade2018data} or by various machine learning models for multi-class classification (e.g., logistic regression (LR), decision tree (DT), and artificial neural network (ANN) \cite{SONG2019464}). In the second step, the models discussed in Section~\ref{subsec: ignoreFM} are used to predict the RUL for each failure mode. In the third step, the final RUL prediction is then calculated as a weighted sum of the RUL predictions for each failure mode. 

Existing works can be categorized into two types based on how they conduct the first and second steps. The first type of model treats the first and second steps as distinct tasks and addresses them sequentially. In particular, \citet{chehade2018data} proposed a composite failure mode index by first combining multiple sensor data to estimate the failure mode probability. The authors then applied quadratic models to estimate the RUL under each failure mode. Similarly, \citet{SONG2019464} employed a neural network to estimate the failure mode probability under a supervised classification framework. Subsequently, the RUL under each failure mode was predicted using another neural network. Although sequentially trained models are generally less complex and easy to train, these models are susceptible to modeling bias and may result in low generalization performance \cite{li2022deep}. 

To overcome these limitations, the second model type combines the first and second steps into a single joint training process. Due to the joint training procedure, these models can well capture dependencies among sensor signals, failure modes, and RUL. As a result, joint models demonstrate superior results in failure mode and RUL prediction than sequentially trained models. For instance, \citet{li2022deep} and \citet{wang2023joint} introduced a deep-branched network for simultaneous failure mode classification and RUL prediction. This approach involves shared feature extraction layers for both tasks with separate sub-networks for supervised failure mode-specific RUL prediction. 

However, all of the aforementioned methods assume that the failure modes of training units are known in advance. In other words, the failure mode predictors are trained under a supervised framework. This assumption is unrealistic since failure mode information is rarely available in practice. How to effectively learn the failure modes of training units when such information is unavailable is still challenging and has yet to be discussed in the literature. This challenge arises from the fact that different sensor signals may exhibit distinct sensor-to-noise ratios and have varying relationships with the underlying failure mechanisms. Moreover, the lengths of the sensor signals differ for each training unit, posing additional challenges for diagnosing the failure modes.

\section{Methodology}\label{sec:method}
\begin{figure*}[htbp]
    \centering
    \includegraphics[width=18cm]{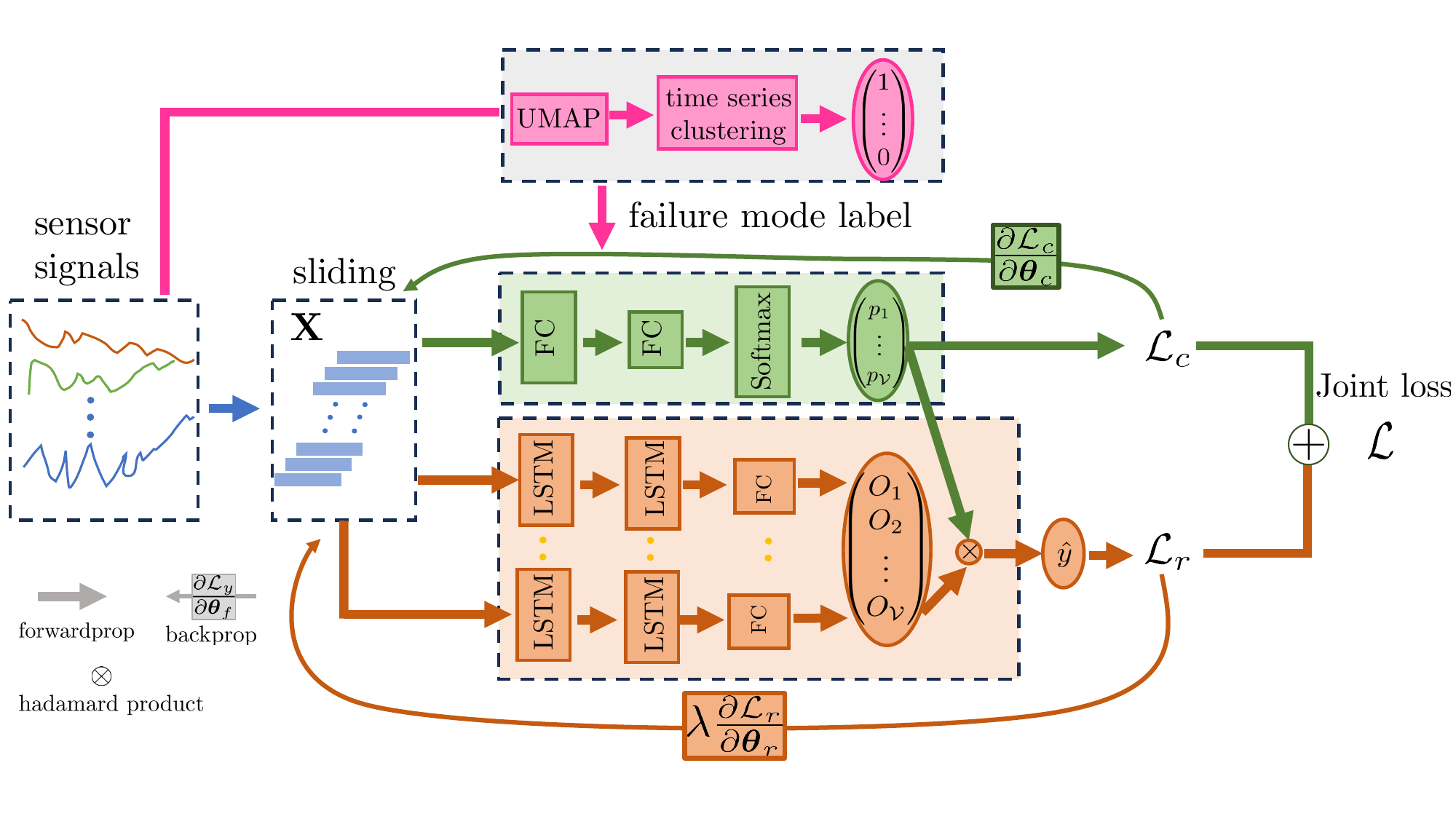}
    \caption{The proposed architecture is composed of three components: the failure mode identifier (red), failure mode predictor (green), and RUL predictor (orange). The high-dimensional sensor signals from $n$ training units are initially processed through UMAP into low-dimensional representations regarding the trajectories of each unit. Then, we employ time series clustering to obtain the failure mode of each training unit. 
    The training unit's sensor signals are first processed via a sliding window method and then simultaneously plugged into both a feed-forward failure mode predictor (green) and an RUL predictor (orange). The joint loss, which combines failure mode classification loss and RUL prediction loss at each iteration, is backpropagated to update the neural network weights.}
    \label{fig:outline}
\end{figure*}

Suppose we have multiple sensors installed in a system collecting condition-monitoring data. Let $x_{i,s,t}$ denote the data of sensor $s$ for unit $i$ at cycle time $t$. Note that $\mathbf{x}_{i,t}\in \mathbb{R}^{S}$ is a vector of measurements of $S$ sensors for unit $i$ at time $t$, where $t=1,2,\dots, T_i$ and $T_i$ is the number of available observations for unit $i$. 
The primary challenges addressed in this study are: 1) identifying potential failure modes of $n$ training units based on the sensor measurements; 2) jointly predicting the failure mode and RUL of the new test units in the field; and 3) incorporating prior domain knowledge (i.e., monotonic constraints) into RUL prediction of the test units. Figure~\ref{fig:outline} illustrates the framework of the proposed method.

\subsection{Failure mode identification/diagnosis for training units}

Training units' failure modes and working conditions are often unavailable in practice. Due to the high dimensionality and complex relations across multiple sensor signals, it is challenging to identify the failure modes directly. To address this issue, we propose a powerful unsupervised approach to automatically identify the failure modes of the training units and discover latent patterns in the sensor signals. In particular, we employ a dimension reduction technique called UMAP \cite{mcinnes2018umap} to gain insights into the underlying degradation process. While UMAP has been heavily used in biological applications, such as analyzing RNA sequence \cite{becht2019dimensionality} and genetics \cite{diaz2021review}, it is not suitable to be directly used for the degradation process as the underlying degradation status continuously evolves over time. To tackle this issue, we characterize the degradation trajectories by plotting them against cycle time after obtaining the low-dimensional representations of the multisensor signals. Finally, we leverage a time series clustering technique to cluster the training units based on their degradation trajectories. Here, each trajectory cluster corresponds to one failure mode. The details of each modeling component are shown below. 

\subsubsection{Dimension Reduction via UMAP}\label{sec: DRviaUMAP}

UMAP is a widely used nonlinear dimension reduction algorithm in the machine learning community. UMAP assumes that the entire data set is uniformly distributed among a topological space (manifold), which can be approximated by finite training samples and projected to a low-dimensional space. The low-dimensional UMAP representations preserve the neighborhood relationships between the data points, making it a valuable tool for visualizing and understanding large, high-dimensional datasets. Existing studies have shown that UMAP can provide state-of-the-art performance with better computational efficiency and better preservation of the local and global data structures compared to other dimension reduction methods such as PCA \cite{pearson1901liii,hotelling1933pca}, Isomap \cite{tenenbaum2000global}, and t-SNE \cite{van2008visualizing}.  While UMAP has been very popular in biological fields, to the best of our knowledge, it has yet to be employed or studied for prognostic applications. 

In particular, let $\mathcal{X} = \{\mathbf{x}_{1,1}, \dots,\mathbf{x}_{i,t}, \dots, \mathbf{x}_{n, T_n}\}$ represent the sensor signals of all training units. The goal of dimension reduction is to find a low-dimensional embedding $\mathcal{Y} = \{\mathbf{y}_{1,1}, \dots, \mathbf{y}_{i,t},\dots, \mathbf{y}_{n, T_n}\}$, where $\mathbf{y}_{i,t} \in \mathbb{R}^D$ and $D<S$ (i.e., $\mathbf{y}_{i,t}$ is the low-dimensional representation of $\mathbf{x}_{i,t}$).

UMAP uses graph layout algorithms to project the high-dimensional data into the low-dimensional space. Specifically, UMAP consists of two primary phases: graph construction in the high-dimensional space and optimization of the low-dimensional graph layout. Details of each phase are as follows:

\textit{i) Graph construction in the high-dimensional space.} For each $\mathbf{x}_{i,t}$, we aim to find its $k$ nearest neighbors $\mathcal{N}(\mathbf{x}_{i,t})$ using a distance metric $\mathrm{d}(\cdot)$. Let $\rho_{i,t}$ denote the minimum positive distance from $\mathbf{x}_{i,t}$ to its neighbors (excluding itself), 
    \begin{align*}
        \rho_{i,t} = \min \{&\mathrm{d}(\mathbf{x}_{i,t},\mathbf{x}_{j,m}) \mid \mathrm{d}(\mathbf{x}_{i,t},\mathbf{x}_{j,m}) >0,  \\
        &\forall \ \mathbf{x}_{j,m} \in \mathcal{N}(\mathbf{x}_{i,t}) \}.
    \end{align*}

Define a weighted graph $\bar{G}(V,E,\mathbf{W})$, with vertices/nodes $V = \mathcal{X}$ and edges (i.e., connections between the nodes) $E = \{ (\mathbf{x}_{i,t},\mathbf{x}_{j,m}): \forall \mathbf{x}_{j,m} \in \mathcal{N}(\mathbf{x}_{i,t}), \forall  \mathbf{x}_{i,t}\in V \}$. The weight on each edge is calculated as follows:
\begin{align*}
    w\left(\left(\mathbf{x}_{i,t}, \mathbf{x}_{j,m}\right)\right) = \exp\left(\frac{-\max\left\{0, \mathrm{d}\left(\mathbf{x}_{i,t}, \mathbf{x}_{j,m}\right) - \rho_{i,t} \right\}}{\sigma_{i,t}}\right).
\end{align*}
    
The weight $w$ here decays exponentially with respect to the distance between $\mathbf{x}_{i,t}$ and its neighbor $\mathbf{x}_{j,m}$. $w$ can be interpreted as the measure of similarity between $\mathbf{x}_{i,t}$ and its neighbor $\mathbf{x}_{j,m}$. In other words, the closer the two are, the higher the weight. The normalization factor $\sigma_{i,t}$ can be computed by binary search such that for a given $\mathbf{x}_{i,t}$ and a fixed number of neighbors $k$, the sum of weights among all its neighbors is a constant value as shown below:    
\begin{align*}
    \sum_{\mathbf{x}_{j,m} \in \mathcal{N}(\mathbf{x}_{i,t})} w\left(\left(\mathbf{x}_{i,t}, \mathbf{x}_{j,m}\right)\right) = \log_2(k).
\end{align*}
The number of neighbors $k$ is one of the most critical parameters in UMAP. In particular, $k$ essentially controls how UMAP balances the local versus global structure. Smaller values of $k$ will push UMAP to focus more on the local structure, while higher values of $k$ prompt UMAP to emphasize the global structure more. In general, this parameter can be tuned by grid search. In our prognostic application, a larger value of $k$ is beneficial since we need to focus more on the global structure to discover useful potential failure mode patterns.

Finally, we convert $\bar{G}(V,E,\mathbf{W})$ into a symmetric, undirected graph $G(V,E,\mathbf{A})$ with a new adjacency matrix $\mathbf{A}$, 
\begin{align*}
    \mathbf{A} = \mathbf{W} + \mathbf{W}^T - \mathbf{W} \circ \mathbf{W}^T,
\end{align*}
where $\circ$ is Hadamard product. This conversion ensures the weight matrix is valid (i.e., symmetric).

\textit{ii) Optimization of the low-dimensional graph layout}. After constructing the high-dimensional graph, UMAP creates a similar graph in the low-dimensional space and optimizes its layout to resemble the high-dimensional structure closely. Denote $H(V', E', \mathbf{B})$ be the weighted graph in the low-dimensional space, with nodes $V' = \mathcal{Y}$ and edges $E = \{ (\mathbf{y}_{i,t},\mathbf{y}_{j,m}): \mathbf{y}_{j,m} \in \mathcal{N}(\mathbf{y}_{i,t}), \forall \mathbf{y}_{i,t}\in V' \}$. Naturally, the weight of edges in $H$ should have a similar form to those in the high-dimension graph $G$, i.e., 
    \begin{align*}
    b\left(\left(\mathbf{y}_{i,t}, \mathbf{y}_{j,m} \right)\right) = \exp \left( \frac{-\max\left\{0, \mathrm{d}\left(\mathbf{y}_{i,t}, \mathbf{y}_{j,m}\right) - \rho^h_{i,t}\right\}}{\sigma_{i,t}^h} \right).
    \end{align*}

Here, we aim to position points $\mathbf{y}_{i,t} \in \mathcal{Y}$ such that the low-dimensional weighted graph $H$ best approximates the high-dimensional weighted graph $G$. In particular, we aim to minimize the dissimilarity (i.e., cross-entropy) between the two graphs by finding the optimal weights $b\left(\left(\mathbf{y}_{i,t}, \mathbf{y}_{j,m} \right)\right)$ that lead to the ideal position of the points $\mathbf{y}_{i,t}$.

However, there are some adjustments needed for the weight $b\left(\left(\mathbf{y}_{i,t}, \mathbf{y}_{j,m}\right)\right)$. Firstly, 
$\rho^h_{i,t}$ is unknown in the low-dimensional graph $H$. In the high-dimensional graph $G$, $\rho_{i,t}$ represents the minimal positive distance from $\mathbf{x}_{i,t}$ to its neighbors, and $\rho^h_{i,t}$ is the low-dimensional equivalent of $\rho_{i,t}$ in $H$. Since $\rho^h_{i,t}$ is unknown, a common practice is to substitute it with a user-specified parameter $\textit{{min\_dist}}$, which represents the minimum distance between all points. Here, $\textit{{min\_dist}}$  essentially controls how closely points are grouped in $H$. Lower values result in more tightly packed embeddings, while larger values encourage a looser arrangement, preserving broader structures. In practice, we experiment with different $\textit{{min\_dist}}$ values to find the optimal value that closely resembles the original graph structure.  Secondly, the normalization factor $\sigma_{i,t}^h$ is often set to $1$. This is possible in $H$ since we can adjust $\mathcal{Y}$ to achieve uniform density. As a result, the weight between $\mathbf{y}_{i,t}$ and $\mathbf{y}_{j,m}$ becomes:
\begin{align*}
	b\left(\left(\mathbf{y}_{i,t}, \mathbf{y}_{j,m}\right)\right) = \begin{cases}
		1 ,  \quad\text{if } \mathrm{d}(\mathbf{y}_{i,t}, \mathbf{y}_{j,m}) \leq \textit{min\_dist},\\
		\exp \left(-\mathrm{d}\left(\mathbf{y}_{i,t}, \mathbf{y}_{j,m} - \textit{min\_dist}\right)\right),  \text{o/w}.
	\end{cases}            
\end{align*}

Since $b\left(\left(\mathbf{y}_{i,t}, \mathbf{y}_{j,m}\right)\right)$ is not continuous, it is challenging to optimize it directly (i.e. cannot directly use gradient descent). Thus, we need to find a continuous, differentiable approximation $b'\left(\left(\mathbf{y}_{i,t}, \mathbf{y}_{j,m}\right)\right)$ with hyper-parameters $\alpha$ and $\beta$, where $\alpha$ and $\beta$ are chosen by non-linear least squares. Consequently, the weights on each edge in the low-dimensional graph $H$ become:
\begin{align*}
    b'\left(\left(\mathbf{y}_{i,t}, \mathbf{y}_{j,m}\right)\right) = \left(1 + \alpha\left( \mathrm{d}\left(\mathbf{y}_{i,t}, \mathbf{y}_{j,m}\right) \right)^\beta \right)^{-1}.
\end{align*}

Then, the optimization process can be executed by stochastic gradient descent. 

\subsubsection{Time series clustering to identify failure modes}\label{sec:time-series-clustering}

After we obtain the low-dimensional representations of each unit at each cycle time via UMAP, our next step is to identify the failure mode of each training unit. 

% Remove them.
% However, as mentioned in Section~\ref{sec: DRviaUMAP}, two crucial parameters in UMAP are the number of neighbors $k$ specified in the high-dimensional structure and the minimum distance $\mathrm{min\_dist}$ specified in the low-dimensional structure. These parameters significantly impact the outcomes of dimension reduction. 

Given that each failure mode exhibits distinctive time-related characteristics, we can trace the degradation trajectory of each unit in the low-dimensional space.

Denote $\mathbf{P}_i = ({\mathbf{y}_{i,1}, \mathbf{y}_{i,2}, \dots, \mathbf{y}_{i,T_i}})^T$ as the trajectory of unit $i$. The set $\mathcal{P} =\{\mathbf{P}_1, \mathbf{P}_2, \dots, \mathbf{P}_n\}$ includes all trajectories of the training units. Thus, our task transforms into clustering all trajectories (time series) into $\mathcal{V}$ groups (failure modes) such that trajectories within each group are similar (i.e., low within-group distance) while ensuring that the trajectories in different groups are dissimilar (i.e., high between-group distance). For the distance measure, we choose Dynamic Time Warping (DTW) \cite{berndt1994DTW} as DTW can accurately capture the similarities between two sequences even under temporal distortions or different sequence lengths.  Formally, we want to partition all trajectories into $\mathcal{V}$ clusters, $\mathcal{C} = \{C_1, C_2, \ldots, C_{\mathcal{V}}\}$, where each cluster $C_\nu \neq \emptyset \in \mathcal{P}$ and $\cup_{\nu=1}^{\mathcal{V}} C_\nu = \mathcal{P}$, such that, 

\begin{align*}
    \mathcal{C} = \underset{\mathcal{C}}{\mathrm{argmin}} \, \sum_{\nu=1}^{\mathcal{V}} \sum_{\mathbf{P}_i \in C_\nu} \mathrm{DTW}\left(\mathbf{P}_i, \mu\left(C_\nu\right)\right)^2,
\end{align*}
where $\mu(C_\nu) = \frac{1}{|C_\nu|} \sum_{\mathbf{P}_i \in C_\nu} \mathbf{P}_i$ is the mean (central) trajectory of cluster $C_\nu$ and $\mathrm{DTW} (\cdot)$ is the dynamic time warping distance between two trajectories.

As a result, the above problem can be reframed into a K-means-based trajectory clustering task with the following steps: 

\begin{enumerate}
	\item Initialize the number of clusters $\mathcal{V}$ by counting the clusters from the UMAP visualization results;
	\item Assign each trajectory to the closest cluster by the $\mathrm{DTW}$ distance measure.
	\item Update the mean\ (central) trajectory of the clusters based on the newly assigned trajectories.
	\item Iteratively perform steps 2) and 3) until the trajectory assignments remain unchanged or the variation falls beneath a predetermined threshold. Note that we run the algorithm multiple times with different initializations to avoid falling into a local optimum. 
\end{enumerate}

\subsection{Joint training of failure mode predictor and RUL predictor} \label{sec:joint-train}

Given that we have identified the failure modes of all training units via UMAP, we propose jointly training the failure mode predictor and RUL predictor to predict the test unit's failure mode and RUL simultaneously. Note that this approach differs from conventional training methods, in which the test failure mode and RUL of test units are predicted sequentially. To train both predictors jointly, we first preprocess the time series (sensor signals) using a sliding window approach. This is achieved by considering previous time steps (within a pre-defined time window size) as inputs and employing the next time step as an output. Formally, for a training unit $i$ with failure mode $\nu$ and time window size $\mathrm{ntw}$, the \textit{input instance} is defined as $\mathbf{X}^\nu_{i,t} = [\mathbf{x}^\nu_{i,t-\mathrm{ntw}+1}, \dots,\mathbf{x}^\nu_{i,t}]^T$.
%and the corresponding $t$ is defined as \textit{instance time id}. 
The corresponding RUL with respect to this input instance is denoted as $y^\nu_{i,t} \in \mathbb{R}$. Then, let $\mathbf{q}^\nu_{i,t} = e_\nu \in \mathbb{R}^\mathcal{V}$ denote the corresponding failure mode vector, where $e_\nu$ is the unit vector with the $\nu$-th element equal to $1$ and all other elements set to $0$, and $\mathcal{V}$ is the total number of failure modes.

Let $G_c(\cdot;\boldsymbol{\theta}_c)$ be the failure mode predictor with parameter $\boldsymbol{\theta}_c$ and $G_c^{\nu}(\cdot;\boldsymbol{\theta}_c)$ be its $\nu$-th element, which is the predicted probability of belonging to the failure mode $\nu$. Also, let $G_r^\nu(\cdot;\boldsymbol{\theta}_r^\nu)$ be the RUL predictor of the $\nu$-th failure mode, and $G_r(\cdot;\boldsymbol{\theta}_r)$ be the final RUL predictor with parameter $\boldsymbol{\theta}_r$. Then, the final RUL prediction is the weighted sum of the predicted RULs under each failure mode:
\begin{align*}
G_r(\cdot;\boldsymbol{\theta}_r) = \sum_{\nu=1}^{\mathcal{V}} G_r^\nu(\cdot;\boldsymbol{\theta}_r^\nu) G_c^{\nu}(\cdot;\boldsymbol{\theta}_c).
\end{align*}

The failure mode classification loss $\ell_c^{i,t}$ and the RUL prediction loss $\ell_r^{i,t}$ with respect to the input instance $\mathbf{X}^\nu_{i,t}$ are represented as:

\begin{align}
\ell_c^{i,t}\left(\boldsymbol{\theta}_c\right) & =\ell_c\left(G_c\left(\mathbf{X}^\nu_{i,t} ; \boldsymbol{\theta}_c\right), \mathbf{q}^\nu_{i,t}\right), \\
\ell_r^{i,t}\left(\boldsymbol{\theta}_r\right) & =\ell_r\left(G_r\left(\mathbf{X}^\nu_{i,t}; \boldsymbol{\theta}_r\right), y^\nu_{i,t}\right).
\end{align}

Then, the proposed network is trained by optimizing the joint total loss across all $n$ training units and time steps: 
 
\begin{align}
\mathcal{L}(\boldsymbol{\theta}_c,\boldsymbol{\theta}_r) =  \sum_{i=1}^{n}\sum_{t=\mathrm{ntw}}^{T_i} \left(\ell_c^{i,t}\left(\boldsymbol{\theta}_c\right) + \lambda \ell_r^{i,t}\left(\boldsymbol{\theta}_r\right)\right), \label{eq:totalLoss}
\end{align}
where $\lambda\geq 0$ is the tuning parameter to balance the two losses. 

In the training process, the failure mode predictor's classification loss $\ell_c^{i,t}$ is set as the cross-entropy between the true failure mode identified via UMAP (i.e., $\mathbf{q}^\nu_{i,t}$), and the estimated failure mode (i.e., $G_c\left(\mathbf{X}^\nu_{i,t} ; \boldsymbol{\theta}_c\right)$):

\begin{align}	
\ell_c^{i,t}\left(\boldsymbol{\theta}_c\right) = -\sum_{\nu=1}^{\mathcal{V}} \mathbf{q}^\nu_{i,t} \ln\left(G_c\left(\mathbf{X}^\nu_{i,t} ; \boldsymbol{\theta}_c\right)\right). 
\end{align}

For the RUL prediction, we consider the health score (HS) loss \cite{ren2023long}, which assigns more weight to late predictions. Let $\ell_r^{i,t,\nu}\left(\boldsymbol{\theta}_r\right)$ denote the RUL predictor loss defined as such: 
\begin{align}
\ell_r^{i,t}\left(\boldsymbol{\theta}_r\right) = \begin{cases} e^\frac{G_r(\mathbf{X}^\nu_{i,t} ; \boldsymbol{\theta}_r) - y^\nu_{i,t}}{10} - 1,  \text { if  }  G_r(\mathbf{X}^\nu_{i,t}) \geq y^\nu_{i,t},\\
e^\frac{y^\nu_{i,t}- G_r(\mathbf{X}^\nu_{i,t} ; \boldsymbol{\theta}_r)}{13} - 1, \text{ o/w }.\end{cases}	
\end{align}

In the evaluations, we use the stochastic gradient descent to optimize this objective function.

\subsection{Imposing monotonic constraints}\label{sec:mono-constraint}

Prior domain knowledge dictates that the RUL predictions should monotonically decrease over time to reflect the deterioration of the system. In other words, we should penalize a model if it predicts a certain RUL at a given time and then predicts a greater RUL at the next time step. Let $ \frac{\mathrm{d}G_r(\mathbf{X}^k_{i,t})}{\mathrm{d}t}$ be the derivative of RUL prediction at time step $t$ for unit $i$ with respect to time. To impose the monotonic constraint, we constrain this derivative to be a negative value, i.e., $	\frac{\mathrm{d} G_r(\mathbf{X}^{\nu}_{i,t})}{\mathrm{d}t} = \zeta	
$, where $\zeta<0$. Therefore, the total loss of all instances with the monotonic constraint is as follows:
\begin{align}
	\min 
	\quad \mathcal{L}(\boldsymbol{\theta}_c,\boldsymbol{\theta}_r) =  &\sum_{i=1}^{n}\sum_{t=\mathrm{ntw}}^{T_i} \left(\ell_c^{i,t}\left(\boldsymbol{\theta}_c\right) + \lambda \ell_r^{i,t}\left(\boldsymbol{\theta}_r\right)\right),\label{eq:obj}\\
    \text{s.t.} \quad  &\frac{\mathrm{d}G_r(\mathbf{X}^\nu_{i,t})}{\mathrm{d}t} = \zeta \quad \forall i,t, \label{eq:monotonic-hard}	\\
    &\zeta<0 \label{eq:rate}.
\end{align}

However, due to the stochastic nature of the degradation process and the inherent random noise in sensor data, imposing a hard constraint like equation~\eqref{eq:monotonic-hard} and ~\eqref{eq:rate} may overly limit the flexibility of the model. Thus, we relax them as soft constrains that allow $\frac{\mathrm{d}G_r(\mathbf{X}^k_{i,t})}{\mathrm{d}t}$ fall within the range of $[\zeta-a, \zeta+a]$, where $a$ is the tolerance and $0<a<-\zeta$. Further, this soft constrain can be incorporated into the model training procedure by introducing it as a regularization term in the objective function:

\begin{align}
\min \quad \mathcal{L'} &= \mathcal{L}(\boldsymbol{\theta}_c,\boldsymbol{\theta}_r) \nonumber \\
&+  \sum_{i=1}^n\sum_{t=\mathrm{ntw}}^{T_i}\left(  \eta\max\left\{0,  \left\lvert\frac{\mathrm{d}G_r(\mathbf{X}^\nu_{i,t})}{\mathrm{d}t} - \zeta\right\rvert - a\right\} \right), \label{eq:final-obj}
\end{align}
where $\eta$ is a tuning parameter to balance the loss between joint training and the additional monotonic constraints.

\section{Case Study}\label{sec:exp}
In this section, the proposed framework is applied to the turbine engine dataset generated from C-MAPSS \cite{MAPSS}, a commercial software widely used to simulate the degradation processes of turbofan aircraft engines. We carry out extensive experiments on this dataset. The source code for this case study is available at: \url{https://github.com/YingFuu/ProgUnknownFMs}.

\subsection{Overview of Dataset}

Aircraft engines are typical examples of complex modern engineering systems. C-MAPSS is the software that simulates the degradation signals of commercial-grade turbofan aircraft engines. Each engine starts with varying degrees of initial wear and unknown manufacturing variations. The software collects 21 degradation signals and three operational variables (i.e., accounting for six working conditions) at every cycle time. Table~\ref{tbl:sensor-desp} provides a detailed description of sensor signals.

The C-MAPSS dataset contains four sub-datasets: FD001, FD002, FD003, and FD004. The details of each sub-dataset are provided in Table~\ref{tbl:overview-ds}. Each sub-dataset is further divided into a training dataset and a test dataset. The signals for the training units are collected from the beginning until failure, while those for the testing units are truncated at a random point before failure. The training units have a large number of cycle time records and thus show clear degradation trends. On the other hand, the test units have partial degradation trends due to the random truncation.  Our primary goal is to identify the failure modes of both training units and test units as well as to predict the RUL of test units by using the available sensor signal data and inferred failure modes. Note that the actual failure modes for both training and test units are unknown.

In this work, we employ four sub-datasets for the low-dimensional visualization of failure modes and working conditions. To illustrate the performance of RUL prediction under unknown failure modes, we focus on the sub-dataset FD003, characterized by two failure modes and one working condition.

\begin{table}
	\centering
	\caption{Detailed description of sensors}
	\label{tbl:sensor-desp}
	\begin{tabular}{ l l l }
		\toprule 
		\text { Symbol } & \text { Description } & \text { Units } \\
		\midrule
		\text { T2 } & \text { Total temperature at fan inlet } & \textdegree \text{R} \\
		\text { T24 } & \text { Total temperature at LPC outlet } & \textdegree \text{R} \\
		\text { T30 } & \text { Total temperature at HPC outlet } & \textdegree \text{R} \\
		\text { T50 } & \text { Total temperature at LPT outlet } & \textdegree \text{R} \\
		\text { P2 } & \text { Pressure at fan inlet } & \text{psia} \\
		\text { P15 } & \text { Total pressure in bypass-duct } & \text{psia} \\
		\text { P30 } & \text { Total pressure at HPC outlet } & \text{psia} \\
		\text { Nf } & \text { Physical fan speed } & \text{rpm} \\
		\text { Nc } & \text { Physical core speed } & \text{rpm} \\
		\text { epr } & \text { Engine pressure ratio (P50/P2) } & - \\
		\text { Ps30 } & \text { Static pressure at HPC outlet } & \text{psia} \\
		\text { phi } & \text { Ratio of fuel flow to Ps30 } & \text{pps/psi}  \\
		\text { NRf } & \text { Corrected fan speed } & \text{rpm} \\
		\text { NRc } & \text { Corrected core speed } & \text{rpm} \\
		\text { BPR } & \text { Bypass ratio } & - \\
		\text { farB } & \text { Burner fuel-air ratio } & - \\
		\text { htBleed } & \text { Bleed enthalpy } & - \\
		\text { Nf\_dmd } & \text { Demanded fan speed } & \text{rpm} \\
		\text { PCNfR\_dmd } & \text { Demanded corrected fan speed } & \text{rpm} \\
		\text { W31 } & \text { HPT coolant bleed } & \text{lbm/s}  \\
		\text { W32 } & \text { LPT coolant bleed } & \text{lbm/s}\\
		\bottomrule
	\end{tabular}
\end{table}

\begin{table}
	\centering
	\caption{Overview of C-MAPSS dataset}
	\label{tbl:overview-ds}
	\begin{tabular}{ccccc}
		\toprule
		& \multicolumn{4}{c}{Sub-dataset} \\ 
		& FD001  & FD002  & FD003 & FD004 \\ \midrule
		\# Training Units     & 100    & 260    & 100   & 248   \\
		\# Testing Units      & 100    & 259    & 100   & 248   \\
		\# Working Conditions & 1      & 6      & 1     & 6     \\
		\# Failure Modes      & 1      & 1      & 2     & 2     \\
  	\makecell{Minimal Cycle Time\\of Training Units}  & 128      & 128      & 145     & 128     \\
    \makecell{Minimal Cycle Time\\of Test Units}       & 31      & 21      & 38     & 19     \\\bottomrule
	\end{tabular}
\end{table}

\subsection{Data preprocessing} \label{Sec: preprocess}
In the datasets, several non-informative sensors provide little information about the degradation status of the system, and thus, they should be dropped before analysis. In particular, a sensor will be removed if it satisfies any of the following conditions:
\begin{itemize}
	\item a sensor only contains a single value,
    \item a sensor has 50\% or more missing values,
    \item a sensor with an extremely low standard deviation in its measurements (less than 0.01).
\end{itemize}

% label the RUL values.
Next, we construct the RUL labels in the training set by subtracting the signal measurement time from the
recorded failure time, following the work \cite{kim2020bayesian}. Specifically, if unit 1 fails after 192 cycles, its RUL would be 191 at the first cycle, 190 at the second cycle, and so forth. The RUL progressively decreases from the initial cycle to 0 when the failure finally occurs. Note that the RUL of the test data is labeled using the same approach.

\subsection{Visualization of Degradation Using UMAP}

In this section, we visualize the C-MAPSS datasets after UMAP dimension reduction to gain insights into their latent structure. In this numerical study, we choose $k=80$, $\textit{min\_dist} = 1$ based on the grid search, where $k$ refers to the number of nearest neighbors preserved in a high dimensional graph and $\textit{min\_dist}$ is the minimum distance between all points in the constructed low dimensional graph. The other parameter settings are in Appendix~\ref{apdix:para-setting-UMAP}.

\begin{figure}[htbp]
	\centering
	\subfloat[\centering FD002: 1 failure mode, 6 working conditions]{\includegraphics[height=3.5cm]{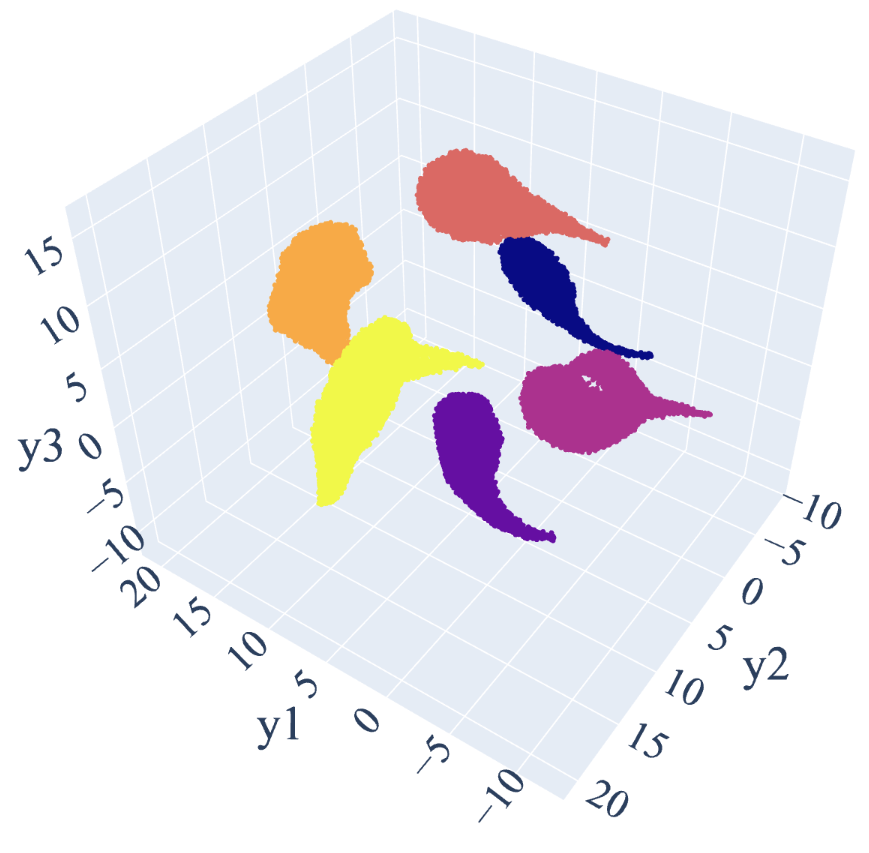} } 
	\qquad
	\subfloat[\centering FD004: 2 failure modes, 6 working conditions]{\includegraphics[height=3.5cm]{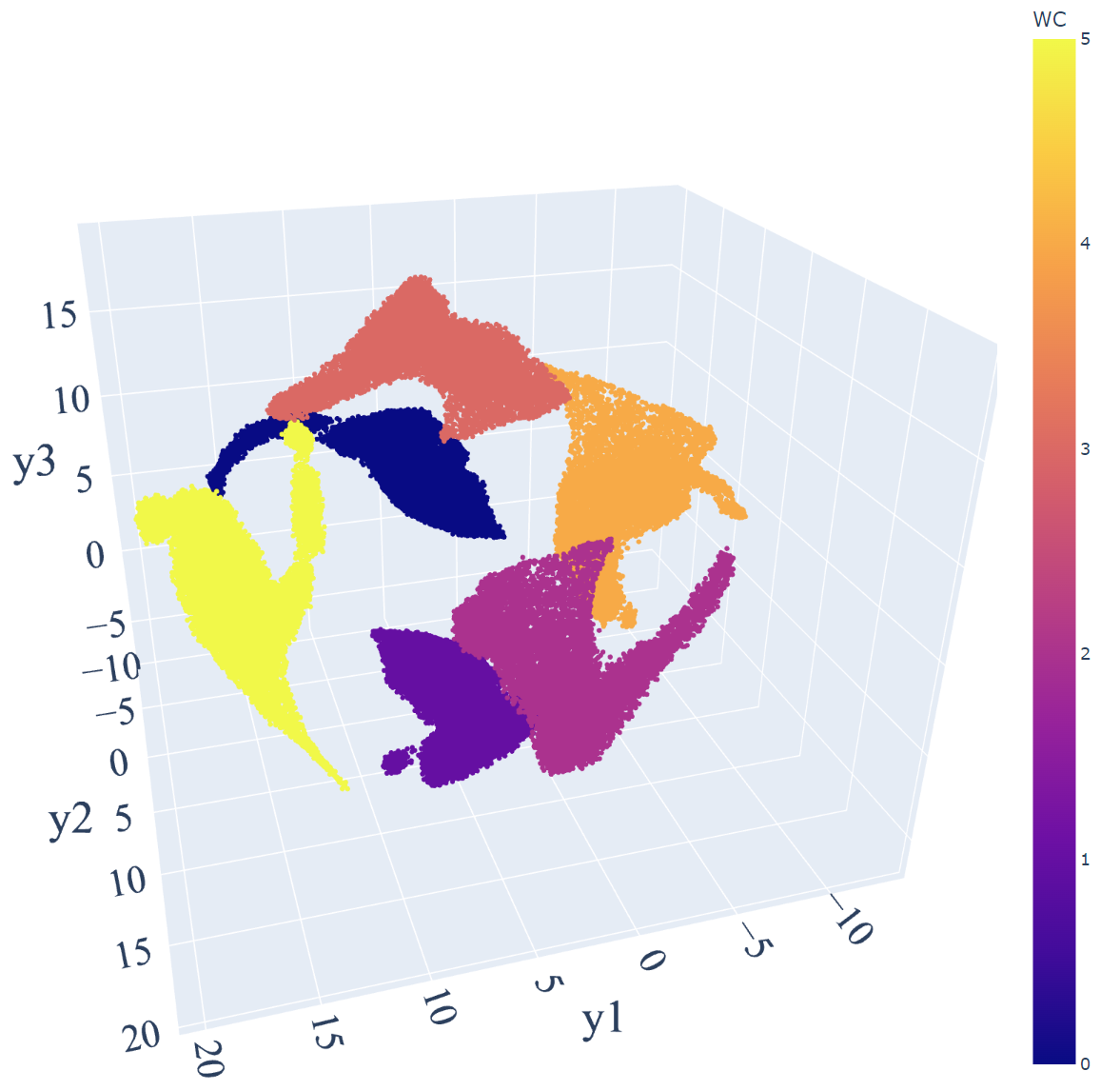} }
	\caption{UMAP projections of C-MAPSS dataset on 3D. Each point represents the three-dimensional representation of each record (one cycle on a unit). Each color corresponds to one working condition.}
    \label{fig:working_condition}
\end{figure}

Figure~\ref{fig: visualization} visualizes the UMAP projections of all four C-MAPSS training datasets in both 2D scatter plots and corresponding 3D line plots (with the RUL as the third dimension). Based on the 2D scatter plots, we have two observations: (i) First, each sub-dataset has one or six clusters, and each cluster corresponds to one working condition. For example, FD001 has only one working condition while FD002 has six working conditions, as detailed in Table~\ref{tbl:overview-ds}. We observe exactly one cluster in Figure~\ref{fig:FD001_2D} and six clusters in Figure~\ref{fig:FD002_2D}. (ii) Second, we assign colors to each point based on its RUL, where blue indicates large RUL values, and orange indicates smaller RUL values. Then, we can clearly see that there are one or two distinct failure trajectories (corresponding to each failure mode) from larger RUL (yellow or blue) to smaller RUL values (red). Specifically, in Figure~\ref{fig:FD001_2D} and Figure~\ref{fig:FD002_2D}, there is only one failure trajectory (i.e., one failure mode), whereas in Figure~\ref{fig:FD003_2D}, there are two failure trajectories (i.e., two failure modes). For the 2D scatter plot of FD004 shown in Figure~\ref{fig:FD004_2D}, there is a slight overlap between clusters due to the effect of multiple failure modes and working conditions. Therefore, we increase the projection dimension to 3D and present a similar 3D scatter plot in Figure~\ref{fig: visualization_3D_scatter_plot} in the Appendix~\ref{sec:3D_vis}. From Figure~\ref{fig: visualization_3D_scatter_plot}, we see that the six clusters are well separated, and two failure trajectories are within each cluster. Moreover, after we color each point according to its respective working condition in the 3D scatter plot, as shown in Figure~\ref{fig:working_condition}, a clear separation between each working condition is observed. This result further confirms that each cluster accurately represents a distinct working condition.

\begin{figure}[htbp]
\centering
    % FD001
     \subfloat[\centering FD001: 2D scatter plot ]{\includegraphics[height=3.2cm]{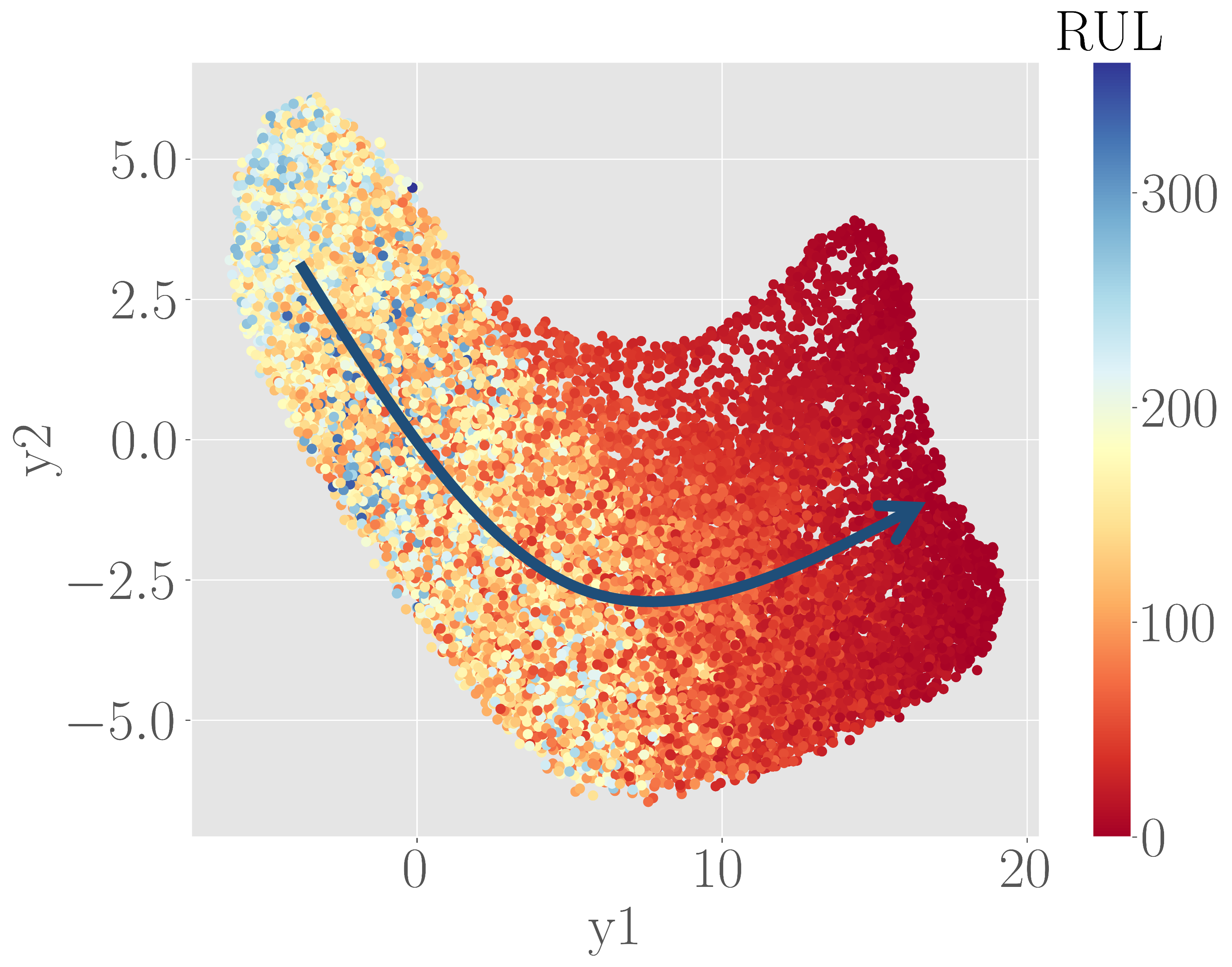}\label{fig:FD001_2D}} 
    \qquad
    \subfloat[\centering FD001: trajectories of all units]{\includegraphics[height=3.2cm]{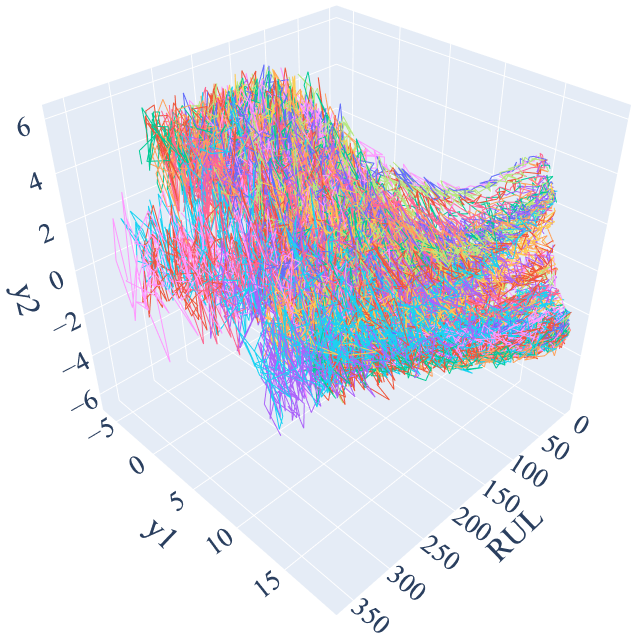}\label{fig:FD001_trajectory}} \\
    
    % FD002
    \subfloat[\centering FD002: 2D scatter plot]{\includegraphics[height=3.2cm]{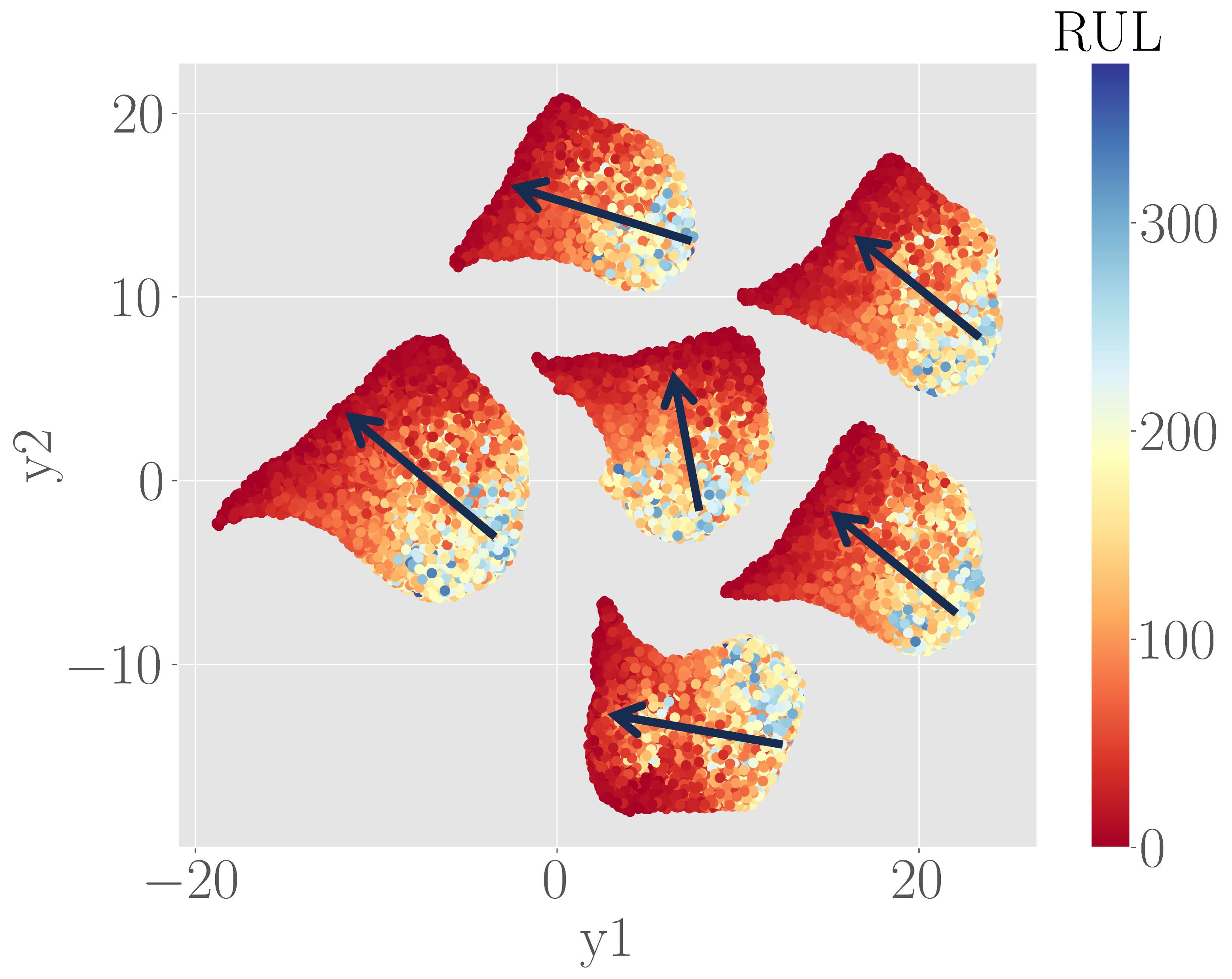}\label{fig:FD002_2D}} 
    \qquad
    \subfloat[\centering FD002: trajectories of all units under the third working condition ]{\includegraphics[height=3.2cm]{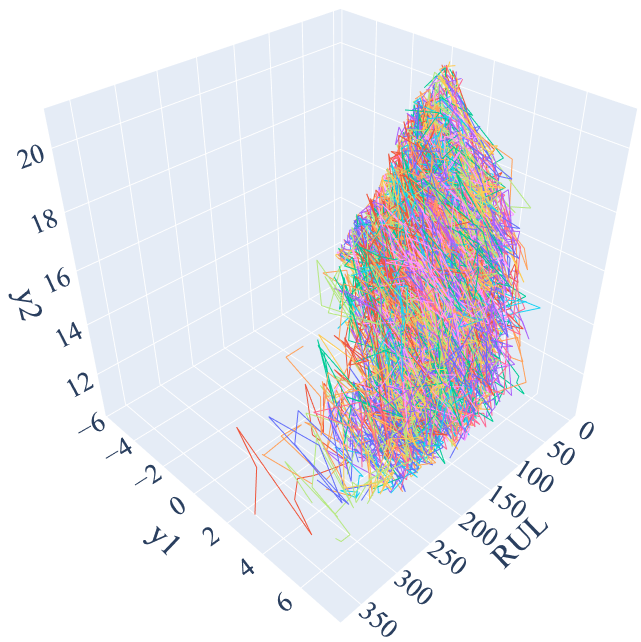}\label{fig:FD002_trajectory}} \\

    % FD003
    \subfloat[\centering FD003: 2D scatter plot ]{\includegraphics[height=3.2cm]{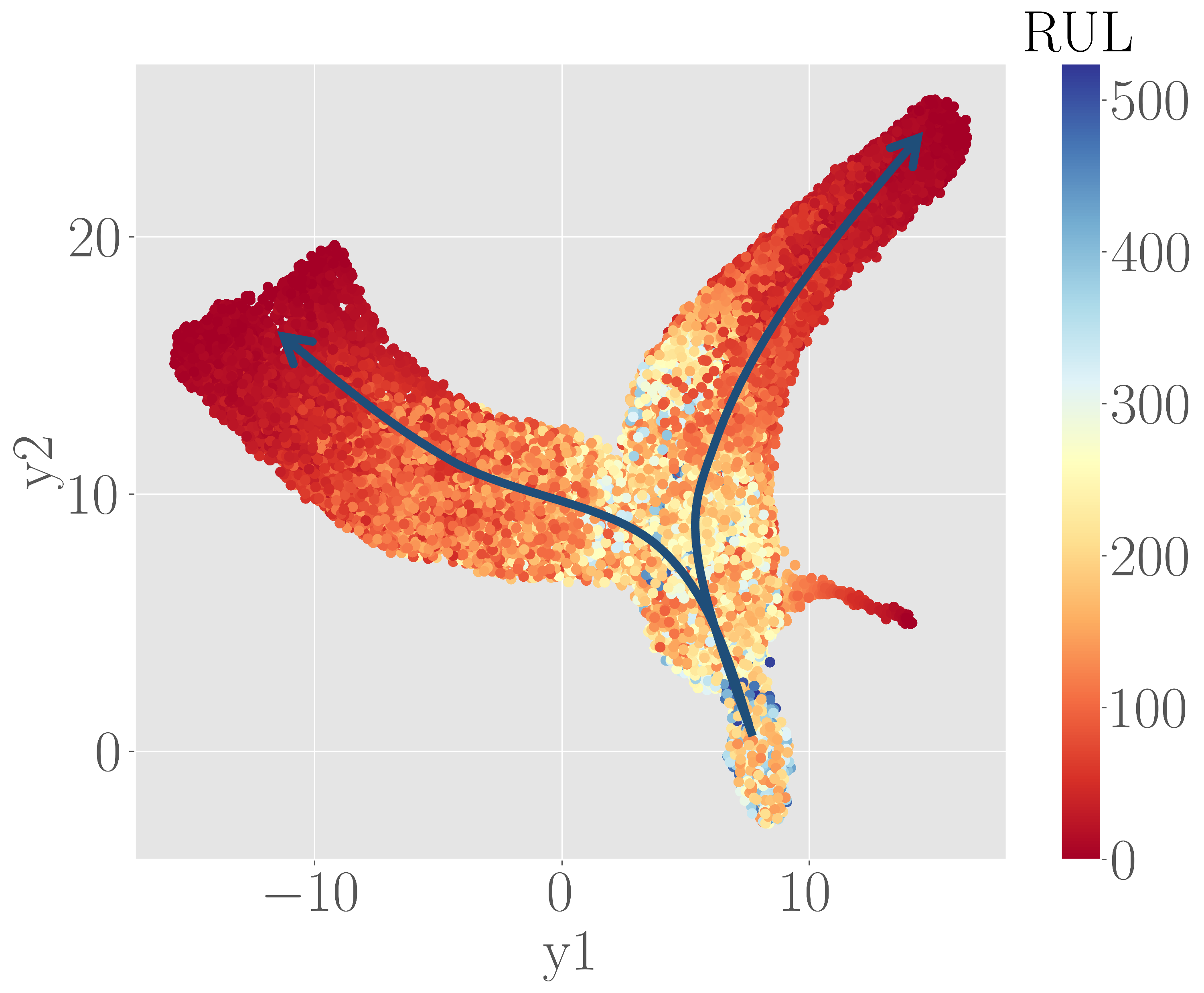}\label{fig:FD003_2D}}
    \qquad
    \subfloat[\centering FD003: trajectories of all units]{\includegraphics[height=3.2cm]{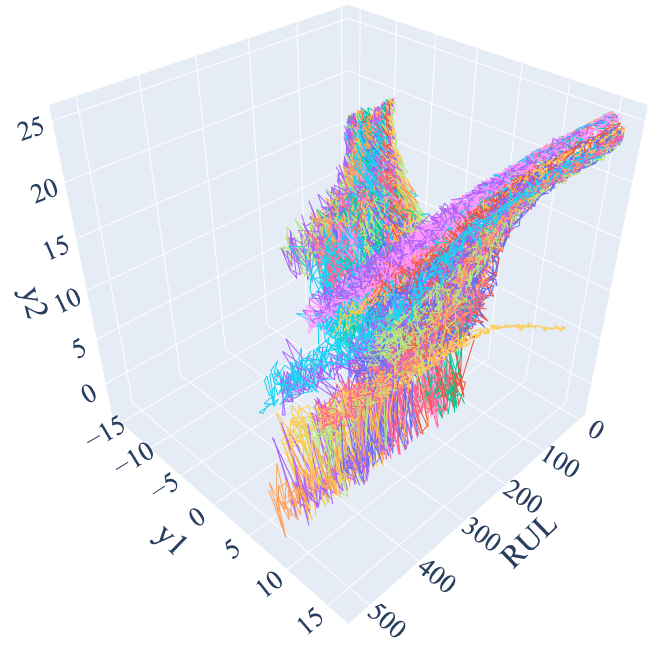}\label{fig:FD003_trajectory}} \\

    % FD004
    \subfloat[\centering FD004: 2D scatter plot]{\includegraphics[height=3.2cm]{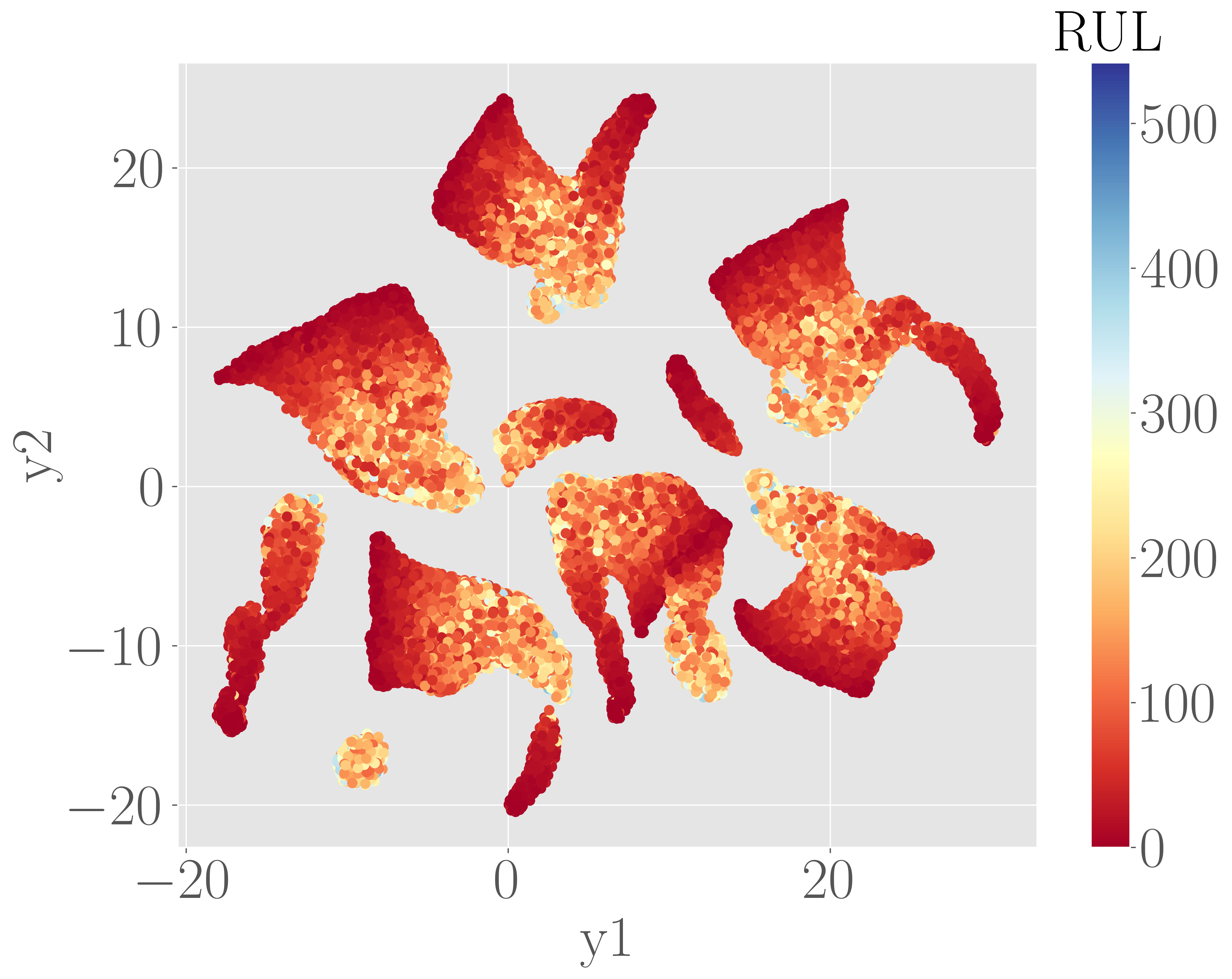}\label{fig:FD004_2D}}
    \qquad
    \subfloat[\centering FD004: trajectories of all units under the third working condition ]{\includegraphics[height=3.2cm]{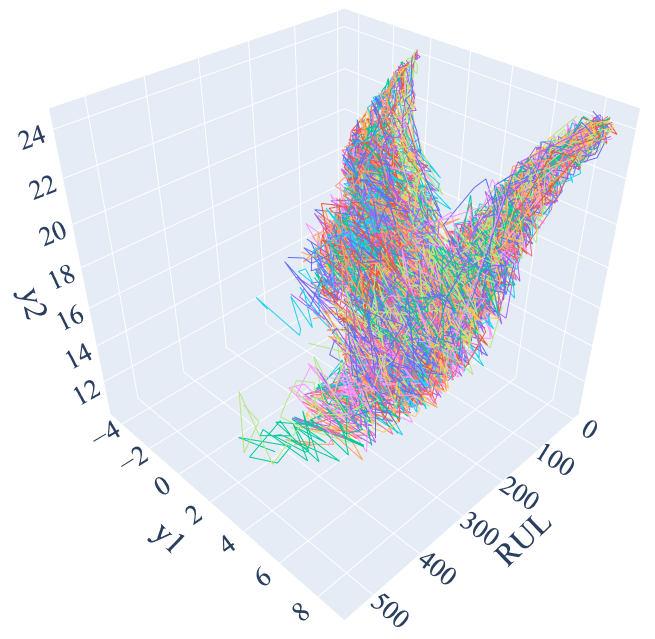}\label{fig:FD004_trajectory}}
 
    \caption{UMAP projections of the C-MAPSS dataset are visualized through 2D scatter plots and corresponding 3D line plots (with the RUL as the third dimension). Within the 2D scatter plots, each point represents the low-dimensional representation (2D) of each record (one cycle on a unit). The colors in the 2D scatter plot indicate the RUL, with blue denoting a greater RUL and orange representing a smaller RUL. Each arrow indicates a failure trend. As for the 3D line plots, each distinct-colored line traces the low-dimensional trajectory of a unit over time.}
    
    \label{fig: visualization} 
\end{figure}

Given that each failure mode exhibits distinctive time-related characteristics, we can also trace the trajectory of each unit in the low-dimensional space by adding the RUL as the third axis. We show the low-dimensional trajectories of all units in  Figure~\ref{fig:FD001_trajectory}, Figure~\ref{fig:FD002_trajectory}, Figure~\ref{fig:FD003_trajectory} and Figure~\ref{fig:FD004_trajectory} for four sub-datasets, respectively. In these figures, each line with a distinct color represents one unit. For datasets with six working conditions (i.e., FD002 and FD004), we illustrate the trajectories under the third working condition. We claim that the failure trajectories across different units reflect the underlying failure modes. To be more specific, for the datasets with a single failure mode (i.e., FD001 and FD002), we observe a singular trend in the failure trajectories of all units. We also identify two separate trends in the failure trajectories in datasets with two failure modes (i.e., FD003 and FD004). This discovery highlights the correspondence between the patterns of trajectories and the multiple failure modes. In particular, we can see that 1) Each trajectory trend represents one failure mode; 2) Each unit is associated with one of the failure modes; and 3) The degradation process of each failure mode is continuous in the low-dimensional feature space. Based on these observations, we then utilize time series clustering techniques to identify the failure mode of each training unit.

\subsection{Results of failure mode identification via time series clustering}

In this section, we show the failure mode identification results of training units via time series clustering.  

\begin{figure}[htbp]
    \centering
    \includegraphics[scale=0.25]{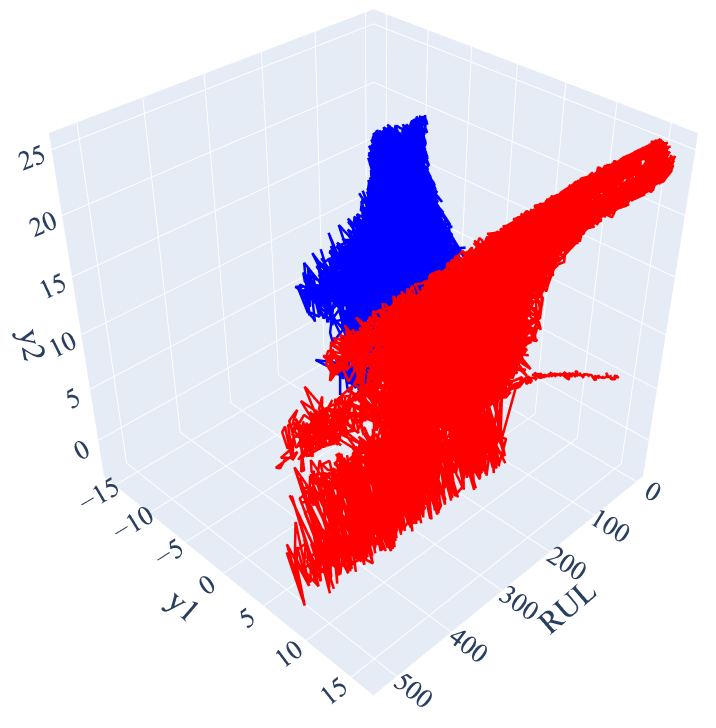}
    \caption{FD003: Trajectories of all units after time series clustering. Each line represents a unit. The color reveals the failure modes.}
    \label{fig:FD003_train_label}
\end{figure}

Figure~\ref{fig:FD003_trajectory} and Figure~\ref{fig:FD003_train_label} show the trajectories of FD003 before and after time series clustering. We can observe that the two failure trajectories are well-separated, with each failure trajectory corresponding to each failure mode. Compared to the previous approaches that assume known failure modes \cite{chehade2018data, wang2023joint}, our approach is more generalizable with fewer assumptions. For instance, in the work by \citet{chehade2018data}, the authors obtained the failure mode information by comparing the trend of six sensor signals heuristically selected between FD001 and FD003. However, this ad-hoc approach is 1) heavily dependent on the reference dataset (i.e., FD001) and 2) unable to detect undocumented failure modes. In contrast, our proposed method is 1) more generalized and less dependent on the reference dataset and 2) able to detect new or unexpected failure modes. Specifically, our method fully leverages the ability of UMAP to capture inherent variability and non-linearity in the data, which makes it even more applicable when the underlying system operates under multiple working conditions.

\begin{figure}[htbp]
    \centering
    \includegraphics[scale=0.2]{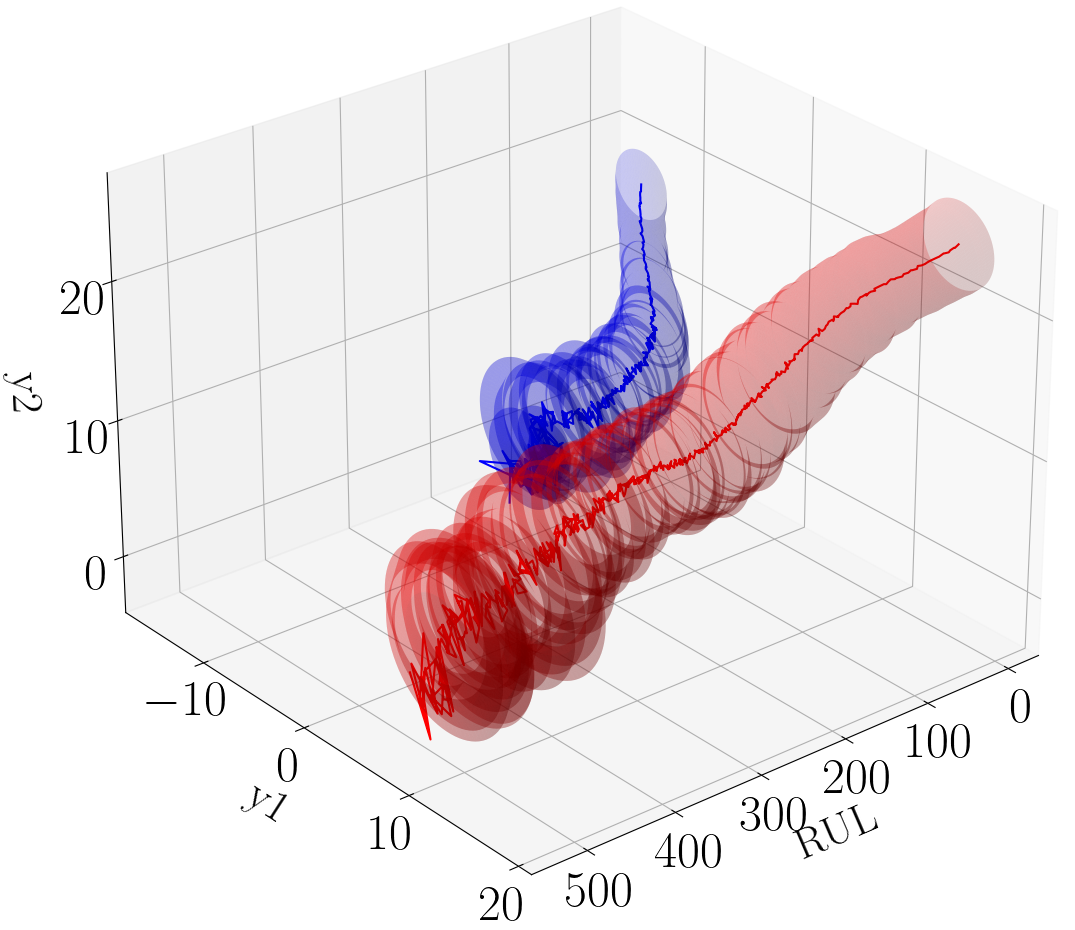}
    \caption{FD003: Mean trajectories of two failure modes. Each color represents one failure mode. The solid line represents the mean trajectory, while the tube represents one standard deviation of trajectories of all units.}
    \label{fig:FD003-tube}
\end{figure}

A fascinating observation arises when we calculate the mean failure trajectories of the training units. In addition, we display one standard deviation around the mean trajectory by plotting a trajectory tube in Figure~\ref{fig:FD003-tube}. Notably, during the initial phase (i.e., larger RUL values), the two mean trajectories are close together, suggesting that differentiating between these failure modes is more challenging. However, the two trajectories diverge as we collect more sensor data and show a clearer separation. Hence, we conclude that the influence of the failure mode becomes more prominent as the unit approaches failure. Further, we also observe that the standard deviation of the trajectories decreases as the units approach failure.

\subsection{RUL prediction}
In this section, we present the RUL prediction results of the proposed joint training framework. Without loss of generality, we employ a feed-forward neural network for the failure mode label predictor and a Long Short-Term Memory (LSTM) network for the RUL predictor with a separate LSTM for each failure mode. We conduct the experiments on FD003, which has two failure modes and one working condition, to demonstrate the RUL prediction results under multiple failure modes. 

First, we apply the min-max normalization to all sensor signals to ensure they fall within the $[0,1]$ range. Then, the sliding time window procedure is applied to all units to augment the training and test dataset. Based on our preliminary investigation and recommendations from relevant works, a window size of $\mathrm{ntw} = 60$ and a stride of $1$ is sufficient to ensure appropriate consideration of long-term dependencies while still maintaining an efficient computational cost. As a result, one instance fed into the neural network contains sensor signals with the recent $60$ time steps. For parameter selection, the best model parameters are found by a grid search with five-fold cross-validation. In each fold, we use 80\% of all training units for training, and the remaining 20\% units are used for validation. The model with the optimal parameters then predicts the RUL of the test units. A detailed description of the parameter selection process can be found in Appendix~\ref{apdix:para-setting-RUL}. 

Since sensor signals are continuously received in a sequence in practice, we examine the whole predicted RUL sequence. The following performance measures are used:

%A common limitation of existing works is that the models are often evaluated based on only the last instance of the test units. For example, consider a test unit with 90 running cycles collected up to a certain point. Setting a window size of 60 with stride 1 results in 31 instances. Then, existing works often evaluate their models based on the predictions made only using the last instance. As a result, such an approach does not fully assess the RUL prediction capability over the course of the degradation. Since sensor signals are continuously received in a sequence in practice, it is important to examine the predicted RUL sequence. Therefore, we evaluate the accuracy of the RUL sequence rather than the ``last snapshot'' of RUL predictions. 

\begin{itemize}
    \item Root Mean Squared Error  (RMSE)
	\begin{align*}
		\mathrm{RMSE} = \sqrt{\frac{1}{|\mathcal{D}|}\sum_{i=1}^n \sum_{t=\mathrm{ntw}}^{T_i} (\hat{y}_{i,t} - y_{i,t})^2}.
	\end{align*}	 
    \item Mean Absolute Error (MAE)
    \begin{align*}
    \mathrm{MAE} = \frac{1}{|\mathcal{D}|}\sum_{i=1}^n \sum_{t=\mathrm{ntw}}^{T_i} |\hat{y}_{i,t} - y_{i,t}|.
	\end{align*}
    
    \item  Mean Absolute Percentage Error (MAPE)

	\begin{align*}
	\mathrm{MAPE} = \frac{1}{|\mathcal{D}|}\sum_{i=1}^n \sum_{t=\mathrm{ntw}}^{T_i} \frac{|\hat{y}_{i,t} - y_{i,t}|}{y_{i,t}}.
	\end{align*}
	
	\item Monotonicity Ratio (MR) 
	\begin{align}
		\begin{gathered}
			\mathrm{MR}=\frac{1}{|\mathcal{D}|}\sum_{i=1}^n \sum_{t=\mathrm{
			ntw}}^{T_i} \mathbf{1}_{\left\{\delta_{i,t}\right\}}, \text { where } \\
			\delta_{i,t}= \begin{cases}1 & \text { if } \frac{d\hat{y}_{i, t}}{dt} < 0, \\
				0 & \text { o/w. }\end{cases}
		\end{gathered}
	\end{align}

\end{itemize}
Here,  $|\mathcal{D}|$ is the total number of instances, and $\hat{y}_{i,t}$ is the estimated RUL, while $y_{i,t}$ is the true RUL. RMSE measures the average magnitude of errors between predicted and actual values. MAE is used to measure the average absolute magnitude of errors. MAPE is used to measure the relative prediction error. Smaller RMSE, MAE, and MAPE values are preferred, indicating better model performance. On the other hand, $\mathrm{MR}$ quantifies the total number of instances with monotonically decreasing RUL predictions, and a larger value is preferable.

\subsubsection{Real-time failure mode probability prediction of test units}

\begin{figure}[htbp]
    \centering
    \includegraphics[scale = 0.25]{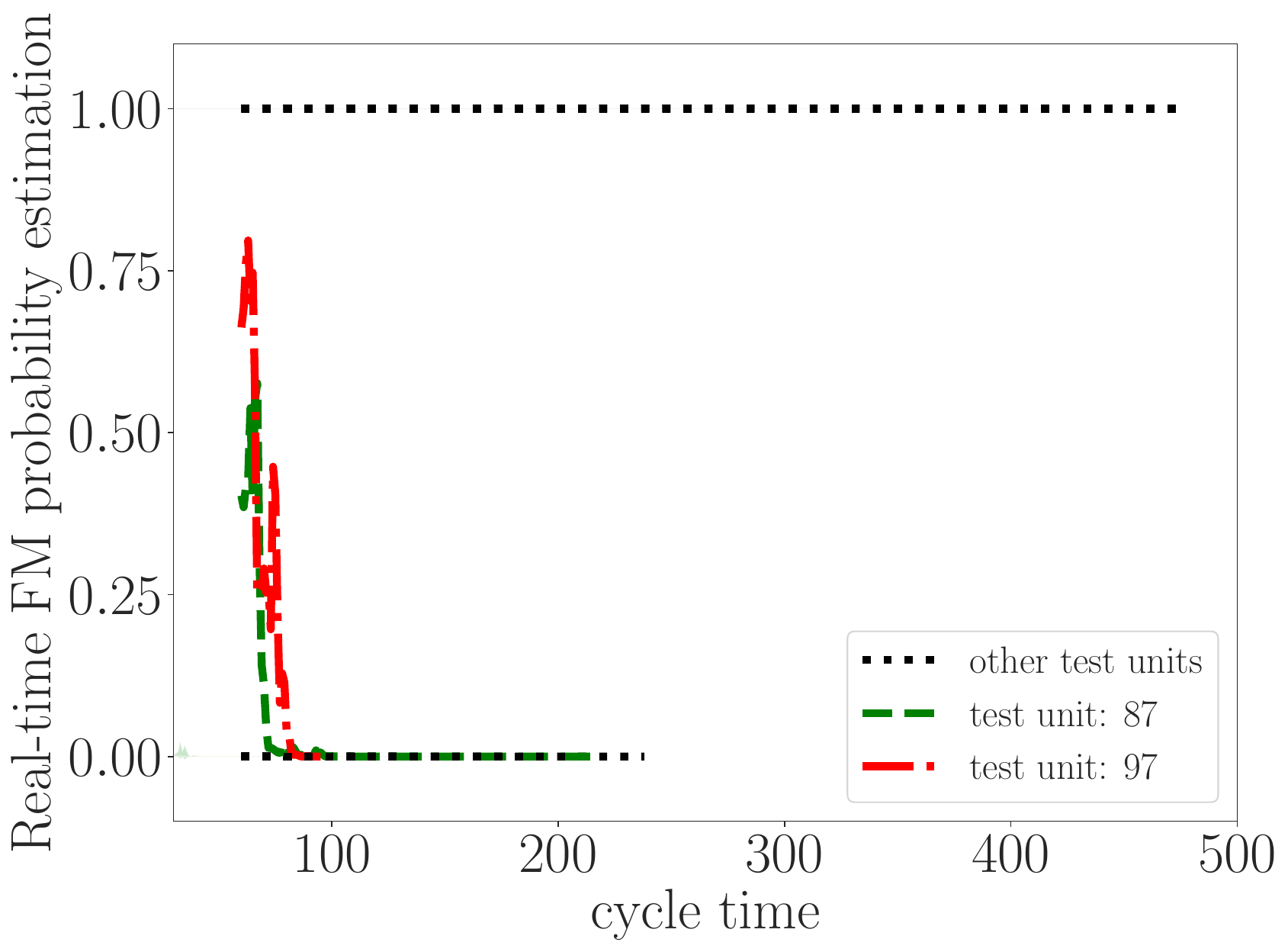}
    \caption{Real-time failure mode probability estimation for FD003 with each line represents one unit. Black dot: other test units; green dash: test unit 87; red dash-dot: test unit 97.}
    \label{fig:prob-estimation}
\end{figure}

This section presents the real-time prediction of failure mode probabilities for test units. Specifically, we illustrate the predicted failure mode probabilities at each cycle time. As previously mentioned, we use a moving window size of $\mathrm{ntw} = 60$ for all input instances in both the failure mode and RUL predictors. The changes in the failure mode probability $P(\nu = 1 \mid \mathbf{X}_{i,t})$ for unit $i$ at instance time $t$ are visualized in Figure~\ref{fig:prob-estimation}. 

It is essential to highlight that our proposed model effectively distinguishes between the two failure modes even from the first input instance (using the initial time window data for each unit) for almost all units except test units 87 and 97. In other words, the failure mode probability rapidly converges to 1 (indicating failure mode 1) or 0 (indicating failure mode 0). Even for units 87 and 97, we observe that the probability estimates for failure mode $0$ (represented as $1 - P(1 \mid \mathbf{X}_{i,t})$) progressively converge to $1$ as we accumulate additional observations. We hypothesize that this initial trend occurs due to the early truncation of signals in the two test units. Nonetheless, with the accumulation of more observations, the proposed model offers a clearer differentiation between the two failure modes.

\subsubsection{Proposed joint training framework v.s. standard framework}
In this section, we compare our proposed joint training framework with a standard framework that ignores failure modes. Note that the proposed method optimizes the joint loss in equation~\eqref{eq:totalLoss} with $\lambda$ as a tuning parameter to balance the loss for failure mode prediction and RUL prediction. The influence of $\lambda$ on RUL prediction error is shown in Table~\ref{tbl:lambda} in the Appendix~\ref{apdix:lambda}.

Figure~\ref{fig:jointLSTMv.s.LSTM} shows the mean absolute error of RUL predictions of the proposed joint LSTM model and the standard LSTM model categorized by RUL intervals (with an interval size of 50). For instance, $0\leq \text{RUL} \leq 50$ provides valuable information for maintenance planning and requires particular attention from the practitioner. Comparing the standard LSTM model with the joint LSTM $(\eta=0)$ model, we observe that the proposed joint LSTM model outperforms the standard LSTM model in terms of average RUL prediction accuracy (smaller absolute error) and lower variance, especially for larger RUL values. This characteristic is highly desired in practice as improving the RUL prediction in the initial stage has been quite challenging but essential for early operation and maintenance planning, ensuring overall cost saving and production efficiency. In addition, knowing failure mode labels in advance not only enhances RUL prediction accuracy but also significantly helps practitioners quickly diagnose and correct the fault.

\begin{figure}[htbp]
	\centering
    \includegraphics[scale = 0.18]{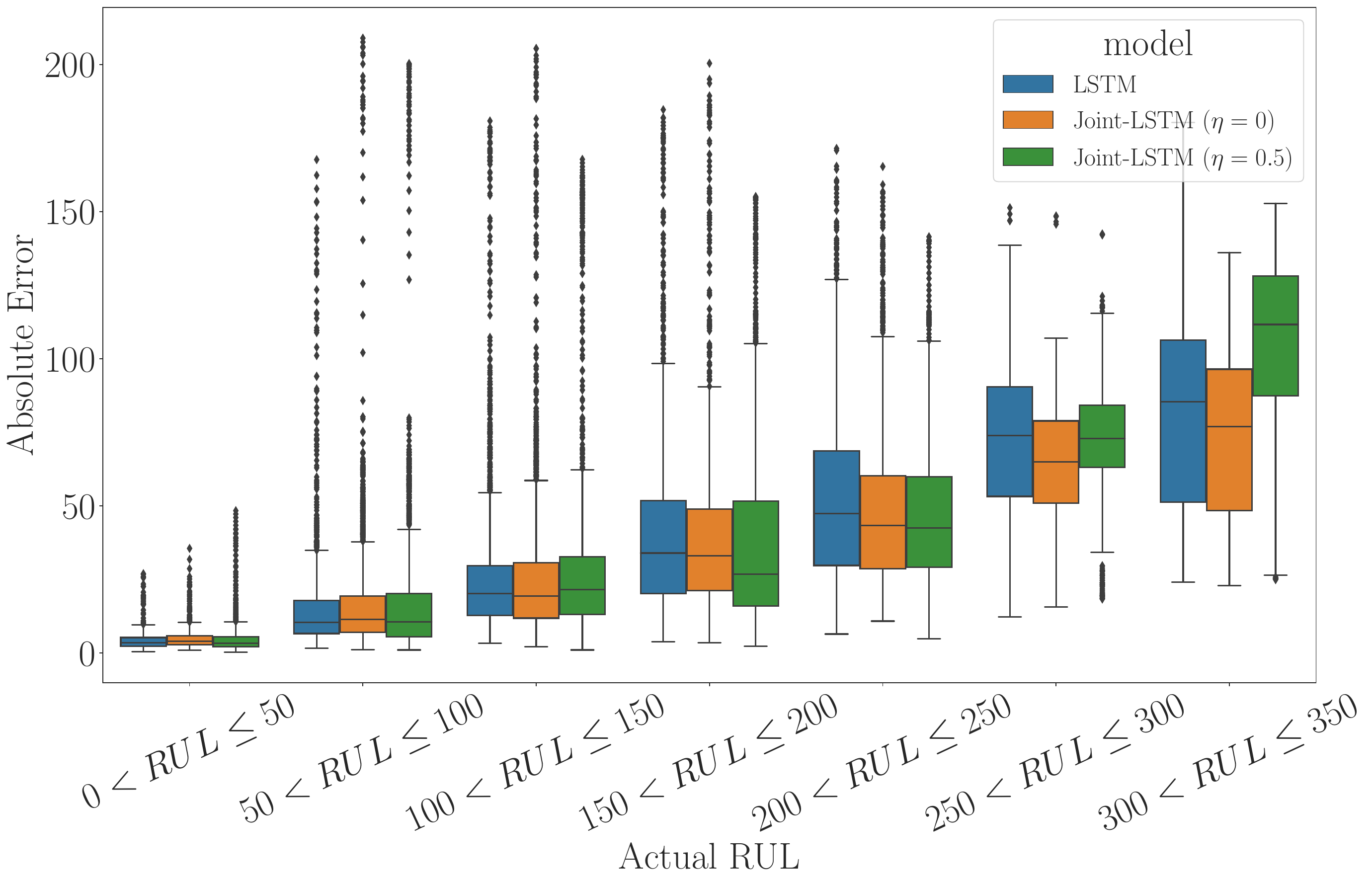}
	\caption{The absolute RUL prediction error of the proposed joint LSTM model with varying monotonic constraints and the standard LSTM model categorized by RUL intervals (with an interval size of 50).}
	\label{fig:jointLSTMv.s.LSTM} 
\end{figure}

\subsubsection{Effect of monotonic constraint}
\begin{figure}[!htbp]
    \centering
    \includegraphics[scale=0.28]{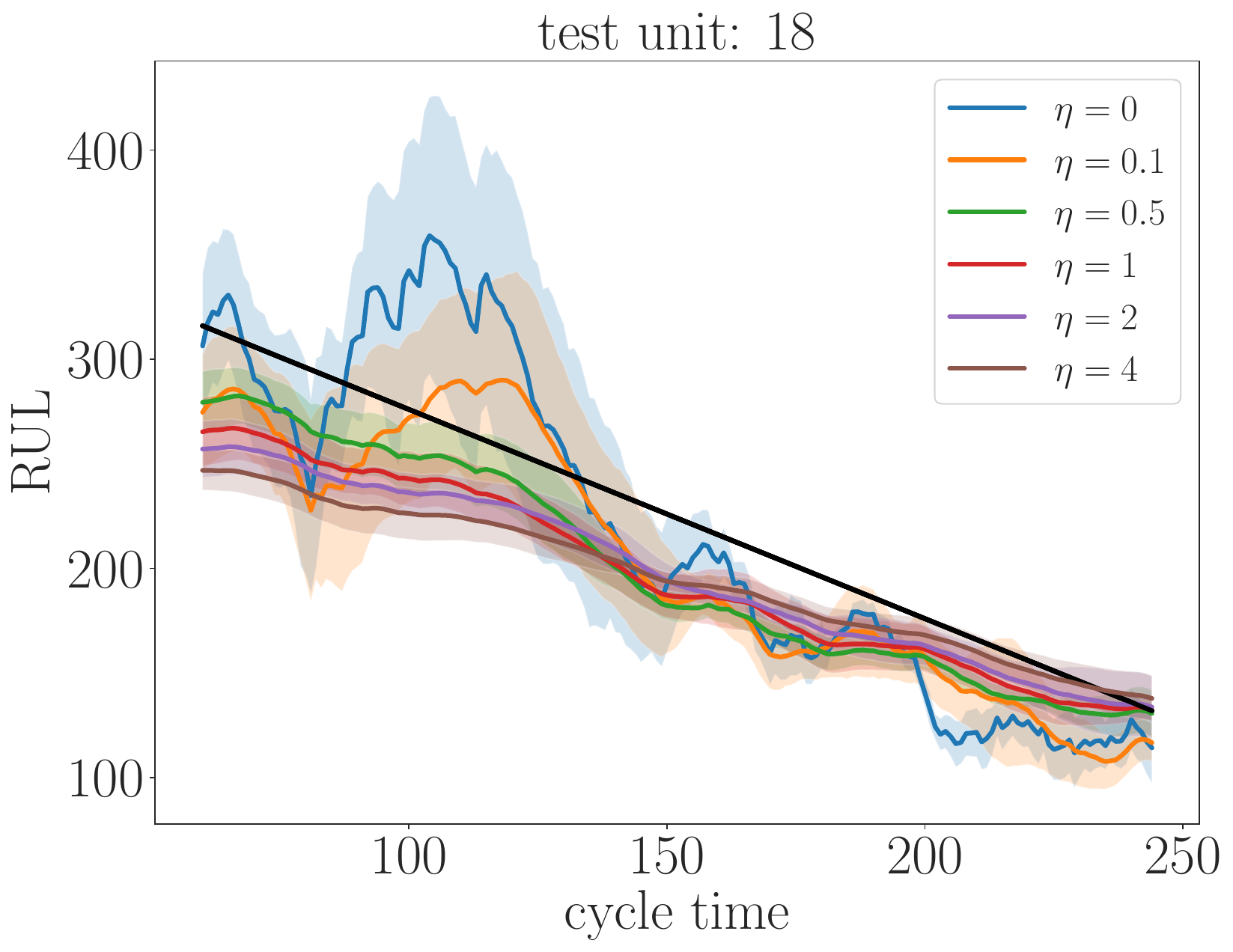}
    \caption{The RUL prediction curves of test unit 18 using the joint-LSTM models with varying monotonicity constraint penalties. The shadowed region indicates the one standard deviation across five-fold cross-validation. The black line is the true RUL.}
    \label{fig:prediction-curve} 
\end{figure}

In this subsection, we will further evaluate the effect of the monotonic constraint. Figure~\ref{fig:prediction-curve} illustrates the RUL prediction curves of one test unit using the joint-LSTM models with varying monotonicity constraint penalties (i.e., $\eta\in \{0.1, 0.5,1,2,4\}$ in equation~\eqref{eq:final-obj}). We can observe that when $\eta=0$, the predicted RUL is not monotonically decreasing with respect to time. As the monotonic penalty $\eta$ increases, the monotonic trend in the RUL predictions becomes stronger. Table~\ref{tbl:rul+monotonicity} gives the RUL prediction results of several benchmark methods and proposed joint LSTM models with monotonic constraints ($\eta$ in equation~\eqref{eq:final-obj}) based on all the available data of the test units. We observe that the overall RUL prediction performance, measured by RMSE/MAPE/MAE, decreases as the monotonic penalty increases. Despite the slight decrease in RMSE/MAPE/MAE, there is a significant rise in the monotonic ratio. This notable gain in MR strengthens the justification for applying the monotonic constraint. Since there is a trade-off between RUL prediction accuracy and monotonicity, we suggest setting the penalty parameter around $\eta = 0.5$ to achieve a balance between accuracy and monotonicity. Specifically, setting $\eta = 0.5$ results in sacrificing approximately 7.15\% of prediction accuracy (from 51.057 to 54.710, measured in terms of RMSE) in exchange for a significant 35.5\% increase in monotonicity (from 0.599 to 0.786, measured in terms of MR). Further increases in the penalty parameter result in diminishing returns in terms of MR over accuracy. As shown in the yellow and green bars in Figure~\ref{fig:jointLSTMv.s.LSTM}, it is worth noting that the decrease in prediction accuracy due to the introduction of monotonicity constraints is more significant in predicting large RUL values (i.e., RUL $>$ 250). However, for RUL values smaller than 250, the joint LSTM model with monotonic constraints begins to outperform. This is expected since model predictions are highly uncertain due to the lack of data during the initial stage of the degradation (i.e., large RUL). As a result, imposing a monotonic constraint damages the model's initial predictive capability. However, as the system collects more data, the prediction uncertainty reduces, and the monotonic constraint starts to provide accurate and interpretable RUL predictions.

Furthermore, we compare the RUL prediction performance with other deep learning-based benchmark methods. With the monotonic constraint $(\eta = 0.5)$, our approach still outperforms CNN and Branch-LSTM \cite{li2022deep} in terms of all the metrics of RMSE, MAPE, MAE, and MR.

\begin{table*}[htbp]
\centering
\begin{threeparttable}
    
\caption{RUL prediction results of benchmark methods and proposed model with monotonic constraints}
\label{tbl:rul+monotonicity}
\begin{tabular}{lllll}
\toprule
Model               & RMSE $\downarrow$           & MAPE $\downarrow$          & MAE $\downarrow$           & MR $\uparrow$           \\ \midrule
CNN \cite{li2018remaining}     & 56.213 (5.409) & 0.244 (0.020) & 37.072 (2.210) & 0.557 (0.004)\\
LSTM \cite{zhang2018long}   & 53.356 (2.608) &\textbf{ 0.217 (0.010)} & 34.940 (1.346) & 0.596 (0.010)\\
Branch-LSTM \cite{li2022deep}   & 59.449 (5.607) & 0.321 (0.069) & 42.634 (6.187) & 0.610 (0.010)\\
Joint-LSTM $(\eta=0)$   & \textbf{51.057 (1.849)} & 0.220 (0.010) & \textbf{34.078 (0.825)} & 0.599 (0.006) \\ 
Joint-LSTM $(\eta=0.1)$ & 54.210 (2.157) & 0.226 (0.014) & 35.553 (1.498) & 0.718 (0.016) \\ 
Joint-LSTM $(\eta=0.5)$ & 54.710 (1.517) & 0.231 (0.007) & 36.381 (0.812) & 0.786 (0.009) \\ 
Joint-LSTM $(\eta=1)$   & 55.859 (2.730) & 0.241 (0.008) & 37.405 (1.340) & 0.812 (0.006) \\ 
Joint-LSTM $(\eta=2)$   & 59.527 (2.937) & 0.256 (0.008) & 39.726 (1.450) & 0.842 (0.008) \\ 
Joint-LSTM $(\eta=4)$   & 60.800 (2.615) & 0.296 (0.008) & 41.953 (1.665) & \textbf{0.8835 (0.014)} \\ \bottomrule
\end{tabular}
\begin{tablenotes}
    \footnotesize
      \item Note: $\eta$ represents the monotonicity constraint penalties in equation~\eqref{eq:final-obj}. The standard deviations are reported in the parenthesis. The down arrow indicates that smaller values are preferable.
    \end{tablenotes}
\end{threeparttable}
\end{table*}

%\subsubsection{Comparision with existing methods}
% Not applicable since they only report the last one. 
%In this subsection, the prognostic performance of the proposed method is compared with the results of the following benchmark methods with respect to the last prediction:
%\begin{enumerate}
%	\item 
%\end{enumerate}

\section{Conclusion} \label{sec:conculsion}
Operating units often experience various failure modes, each leading to distinct degradation paths. Thus, performing failure mode diagnostics is crucial. Relying solely on a single unified prognostic model for RUL prediction may result in sub-optimal performance across different failure modes. Current approaches either ignore failure modes or assume known failure mode labels, which can be challenging to acquire in practice. Additionally, many existing deep learning-based prognostic models ignore the output monotonicity in the context of RUL prediction. To address these challenges, we propose using UMAP to identify failure modes and capture the degradation trajectory of each unit. Subsequently, we design a time series-based clustering method to determine the failure modes of training units. Then, we further introduce a monotonically constrained prognostic model for predicting the failure modes and RUL of test units. Particularly, the predicted failure mode information is incorporated into the final RUL prediction while ensuring that the RUL predictions are monotonic during degradation. Due to the flexibility of the proposed model and its unique advantage in handling unknown failure modes, we expect the proposed method to be widely adopted in various practical scenarios, especially in manufacturing systems with complex structures and unknown failure modes.

While the proposed method is promising, we noticed that for some units subject to complex working conditions, tracking their failure trajectories via UMAP is still challenging due to their tendency to transition randomly across working conditions. This randomness disrupts the continuity of the trajectories, making it difficult to characterize the unit's degradation paths. In future research, we will develop more advanced methods tailored to degradation units under multiple failure modes and diverse working conditions.

% Can use something like this to put references on a page
% by themselves when using endfloat and the captionsoff option.
\ifCLASSOPTIONcaptionsoff
\newpage
\fi

% trigger a \newpage just before the given reference
% number - used to balance the columns on the last page
% adjust value as needed - may need to be readjusted if
% the document is modified later
%\IEEEtriggeratref{8}
% The "triggered" command can be changed if desired:
%\IEEEtriggercmd{\enlargethispage{-5in}}

% references section

% can use a bibliography generated by BibTeX as a .bbl file
% BibTeX documentation can be easily obtained at:
% http://mirror.ctan.org/biblio/bibtex/contrib/doc/
% The IEEEtran BibTeX style support page is at:
% http://www.michaelshell.org/tex/ieeetran/bibtex/
% Generated by IEEEtranN.bst, version: 1.14 (2015/08/26)

%
% <OR> manually copy in the resultant .bbl file
% set second argument of \begin to the number of references
% (used to reserve space for the reference number labels box)
%\begin{thebibliography}{1}
%
%\bibitem{IEEEhowto:kopka}
%H.~Kopka and P.~W. Daly, \emph{A Guide to \LaTeX}, 3rd~ed.\hskip 1em plus
%  0.5em minus 0.4em\relax Harlow, England: Addison-Wesley, 1999.
%
%\end{thebibliography}

% biography section
% 
% If you have an EPS/PDF photo (graphicx package needed) extra braces are
% needed around the contents of the optional argument to biography to prevent
% the LaTeX parser from getting confused when it sees the complicated
% \includegraphics command within an optional argument. (You could create
% your own custom macro containing the \includegraphics command to make things
% simpler here.)

\newpage

\appendices

\section{Experiment Details}
\subsection{Parameter Configuration for UMAP}\label{apdix:para-setting-UMAP}

As we mentioned in the main paper, the number of neighbors used for manifold approximation in a high-dimensional graph, $k$, and the minimum distance between embedded points in a low-dimensional graph and embedded dimensionality, $\textit{{min\_dist}}$, are considered the most critical parameters for UMAP. They are explored by the grid search. For other parameters, the default value is used. Below is the summary of parameter settings involved in UMAP:
\begin{itemize}
    \item \textit{n\_neighbors}: the number of neighbors used for manifold approximation in a high dimensional graph, chosen from the list \{15, 80, 100\}.
    \item \textit{min\_dist}: minimum distance between embedded points in a low-dimensional graph, chosen from \{0.1, 1\}.
    \item \textit{n\_components}: The embedded dimensionality, chosen from \{2,3\}.
    \item \textit{metric = 'euclidean'}: metric to compute distances in high dimensional space.
    \item \textit{init='spectral'}: the method used to initialize the low dimensional embedding. 
    \item \textit{learning\_rate=1.0}: the initial learning rate for the embedding optimization.
\end{itemize}

\subsection{Parameter Configuration for RUL prediction}\label{apdix:para-setting-RUL}
The parameters related to RUL prediction are mainly the ones involved in the neural networks. Below is a list of parameters whose values are chosen by the cross-validation:
\begin{itemize}
    \item network: We tried two hidden layers with various neurons in the hidden layers \{(16, 16), ( 16, 32), (32, 32),(32, 64), (64, 64)\}. To avoid overfitting, a configuration is skipped if the number of trainable weights exceeds 2/3 of the training instances.
    \item activation:  \{`relu'\}.
    \item solver: \{`Adam'\}.
    \item learning\_rate: \{ 0.0001\}.
\end{itemize}

A network was trained for \textit{${max\_iter} = 2000$} epochs. We fixed the seed for pseudo-random number generator \textit{${random\_state} = 42$} so our results can be reproduced. 

All other parameters use the default value. All experiments are implemented in Python 3.9.12, using PyTorch 1.12.0. The experiments were conducted on laptops with an Intel(R) Core(TM) i7-11370H CPU @ 3.30GHz and 32GB RAM and an NVIDIA GeForce RTX 3050 Ti GPU.

\section{Additional Experiments}
\subsection{3D UMAP scatter plot}\label{sec:3D_vis}
\begin{figure}[htbp]
\centering
     \subfloat[\centering FD001]{\includegraphics[height=3.8cm]{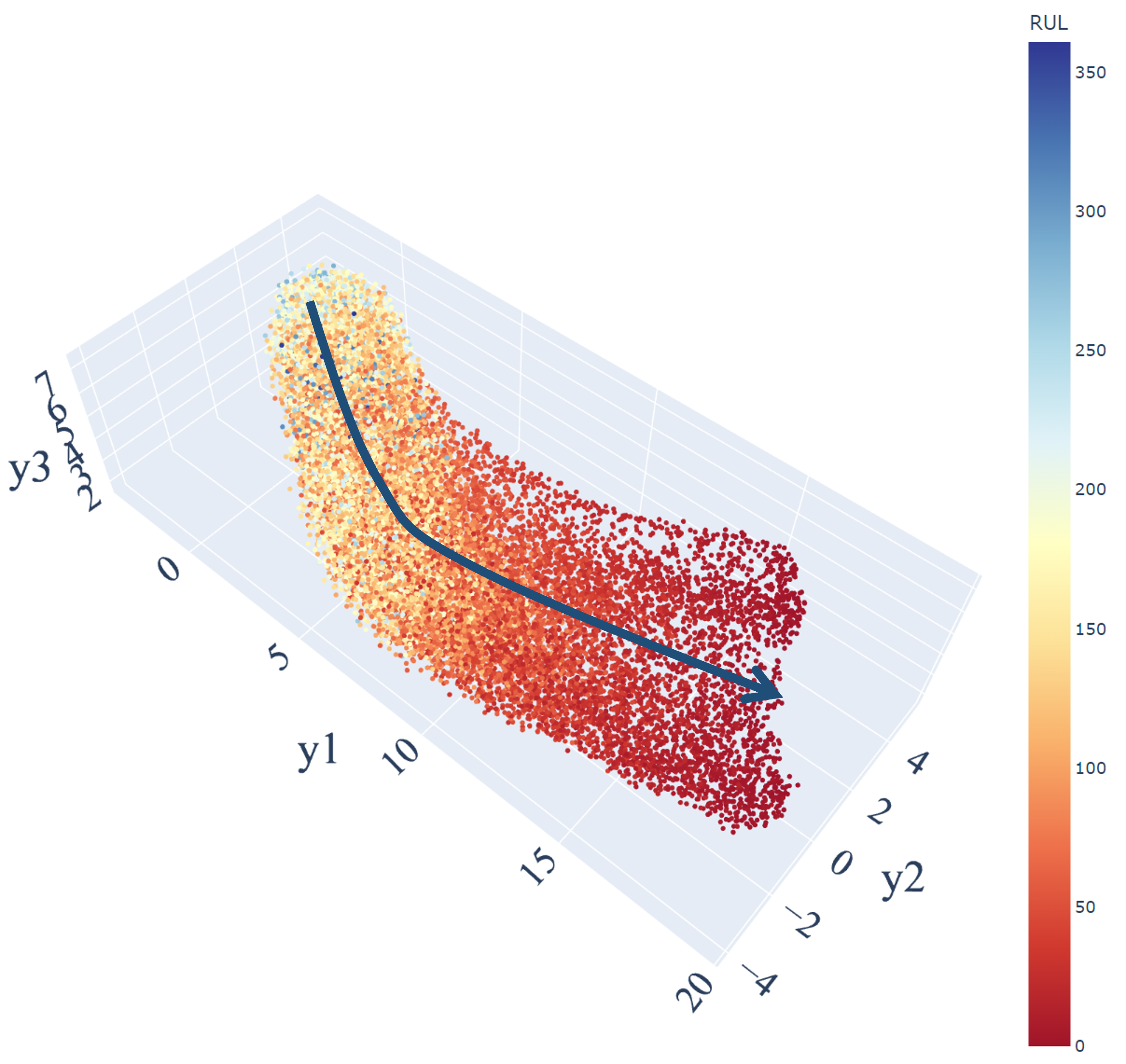}\label{fig:FD001_3D}} 
    \qquad
    \subfloat[\centering FD002]{\includegraphics[height=3.8cm]{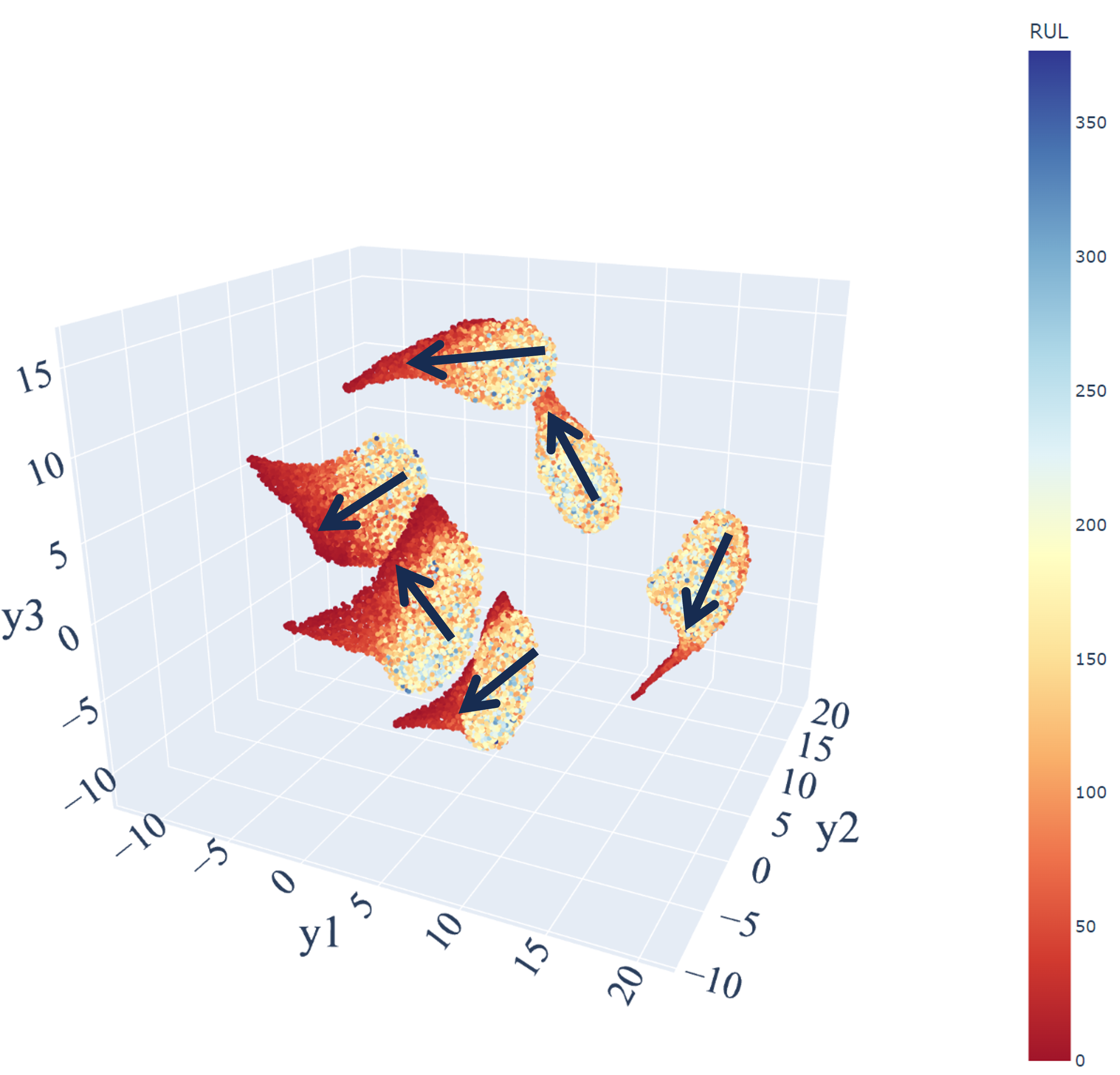}\label{fig:FD002_3D}} \\
    \subfloat[\centering FD003]{\includegraphics[height=3.8cm]{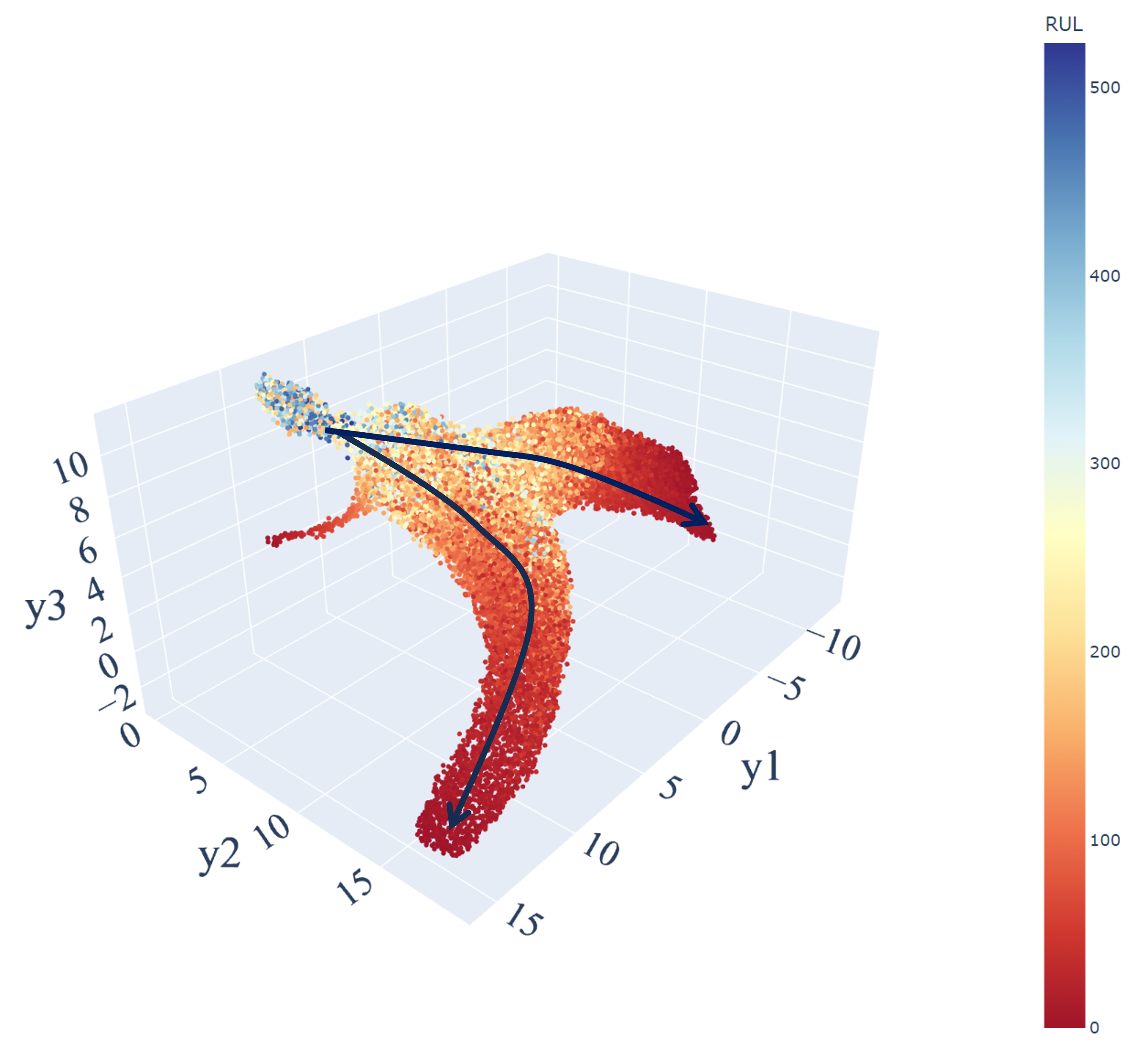}\label{fig:FD003_3D}} 
    \qquad
    \subfloat[\centering FD004]{\includegraphics[height=3.8cm]{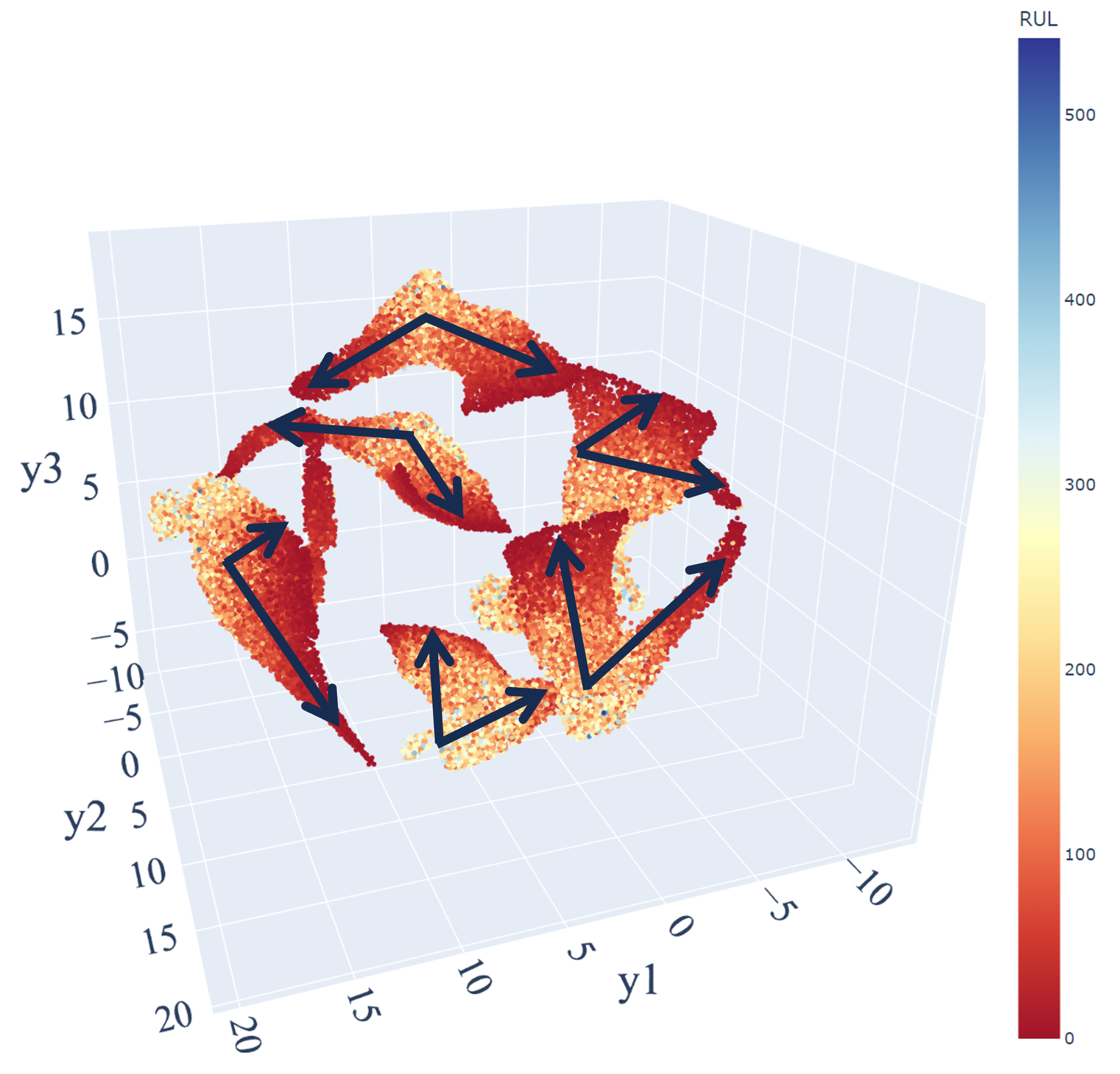}\label{fig:FD004_3D}} 
    \caption{UMAP projections of the C-MAPSS dataset visualized in 3D scatter plots. Within the 3D scatter plots, each point represents the low-dimensional representation (3D) of each record (one cycle on a unit). The colors in the 3D scatter plot indicate the RUL, with blue denoting a greater RUL and orange representing a smaller RUL. The arrow indicates the trend of failure. }
    \label{fig: visualization_3D_scatter_plot} 
\end{figure}

Figure~\ref{fig: visualization_3D_scatter_plot} shows the 3D scatter plot visualization of four sub-datasets within the C-MAPSS dataset. The color represents the RUL, where blue indicates a greater RUL and orange denotes a smaller RUL. Each arrow in the figures illustrates the trend of failure. We have the following observations: i) In datasets with six distinct working conditions, each corresponding to a specific cluster, the clusters exhibit clear separation (see Figure~\ref{fig:FD002_3D} and Figure~\ref{fig:FD004_3D}). ii) For datasets with two failure modes, there are two failure trajectories within each cluster, as illustrated in Figure~\ref{fig:FD003_3D} and Figure~\ref{fig:FD004_3D}.

\subsection{Effect of $\lambda$}\label{apdix:lambda}

Table~\ref{tbl:lambda} illustrates the impact of the tuning parameter $\lambda$ on the prediction performance, which balances the RUL prediction loss and the failure mode classification loss. It shows that increasing $\lambda$ puts a higher emphasis on the RUL prediction loss and less on failure mode prediction. In addition, the overall performance initially decreases but gradually improves as $\lambda$ becomes larger. An interesting observation is that when we only consider the RUL prediction loss (i.e., $\lambda = \inf$), the performance is worse than the joint loss. In other words, jointly optimizing the failure mode classification and the RUL prediction loss results in a more accurate prognostic model. In this study, we set $\lambda = 10$ based on the experiment results.

\begin{table}[htbp]
\centering
\begin{threeparttable}
\caption{RUL prediction results with different $\lambda$}

\label{tbl:lambda}
\begin{tabular}{llll}
\toprule
$\lambda$        & RMSE $\downarrow$           & MAPE $\downarrow$          & MAE $\downarrow$      \\ \midrule
1                & 52.764 (2.676) & 0.217 (0.007) & 34.744 (0.857)  \\ 
5                & 54.584 (1.381) & 0.230 (0.009) & 35.815 (1.016)  \\
10               & \textbf{51.057 (1.849)} & \textbf{0.220 (0.010)} & \textbf{34.078 (0.826)}  \\ 
20               & 53.528 (1.928) & 0.238 (0.034) & 36.114 (1.960)  \\ 
inf\tnote{1}             & 53.288 (4.007) & 0.222 (0.009) & 35.295 (2.210)  \\ 
\bottomrule
\end{tabular}
    \begin{tablenotes}
    \footnotesize
      \item Note: Standard deviations are reported in the parenthesis. The down arrow indicates smaller values are preferable.
      \item[1] Model setting that ignores failure mode classification loss.
    \end{tablenotes}
\end{threeparttable}
\end{table}

\end{document}